\documentclass{article} % For LaTeX2e
\usepackage{collas2023_conference,times}
\usepackage{easyReview}
\usepackage[T1]{fontenc}
\usepackage{cancel}

\usepackage{amsmath,amsfonts,bm}

\def\eqref#1{equation~\ref{#1}}
\def\1{\bm{1}}

\DeclareMathAlphabet{\mathsfit}{\encodingdefault}{\sfdefault}{m}{sl}
\SetMathAlphabet{\mathsfit}{bold}{\encodingdefault}{\sfdefault}{bx}{n}

\usepackage{hyperref}
\hypersetup{
    colorlinks=true,
    linkcolor=red,
    filecolor=magenta,
    urlcolor=blue,
    citecolor=purple,
    pdftitle={Overleaf Example},
    pdfpagemode=FullScreen,
    }

\usepackage{booktabs}
\usepackage{arydshln}
\usepackage{multirow}
\usepackage{caption}
\usepackage{subcaption}
\usepackage{wrapfig}

\title{ Re-Weighted Softmax Cross-Entropy to Control Forgetting in Federated Learning}

\author{Gwen~Legate \\
Concordia University, Mila\\
\texttt{gwendolyne.legate@mila.quebec} \\
 \And
  Lucas~Caccia\\
  McGill University, Mila\\
  \texttt{lucas.page-caccia@mail.mcgill.ca} \\
  \And
  Eugene~Belilovsky\\
  Concordia University, Mila\\
  \texttt{eugene.belilovsky@concordia.ca} \\
}

\preprintcopy % Uncomment for the preprint version, but NOT for submission.

\begin{document}

\maketitle

\begin{abstract}
In Federated Learning, a global model is learned by aggregating model updates computed at a set of independent client nodes, to reduce communication costs multiple gradient steps are performed at each node prior to aggregation. A key challenge in this setting is data heterogeneity across clients resulting in differing local objectives which can lead clients to overly minimize their own local objective, diverging from the global solution. We demonstrate that individual client models experience a catastrophic forgetting with respect to data from other clients and propose an efficient approach that modifies the cross-entropy objective on a per-client basis by re-weighting the softmax logits prior to computing the loss. This approach shields classes outside a client’s label set from abrupt representation change and we empirically demonstrate it can alleviate client forgetting and provide consistent improvements to standard federated learning algorithms. Our method is particularly beneficial under the most challenging federated learning settings where data heterogeneity is high and client participation in each round is low.
\end{abstract}

%%%%%%%%% BODY TEXT
\section{Introduction}\label{sec:intro}
Federated Learning (FL) is a distributed machine learning paradigm in which a shared global model is learned from a decentralized set of data located at a number of independent client nodes \citep{mcmahan2017communication,konevcny2016federated}. Driven by communication constraints, FL algorithms typically perform a number of local gradient update steps before synchronizing with the global model. This reduced communication strategy is very effective under independent and identically distributed (i.i.d.) settings, but data heterogeneity across clients has direct implications on the convergence and performance of FL algorithms \citep{zhao2018federated}. FL was conceptualized as a learning technique to train a shared model without sharing user sensitive data, while allowing users to benefit from data stored at other nodes, such as phones and tablets of decentralized users. Under realistic settings, client data will often have non-i.i.d. distributions. For the case of supervised multi-class classification, users may frequently have no data whatsoever from one or several classes. Data in-homogeneity across clients frequently induces client drift, a phenomenon in which clients progress too far towards optimizing their own local objective, leading to a solution that has severely "drifted" from an optimal global solution \citep{karimireddy2020scaffold}. 

\begin{figure}[t]
    \centering
    \includegraphics[width=0.95\textwidth]{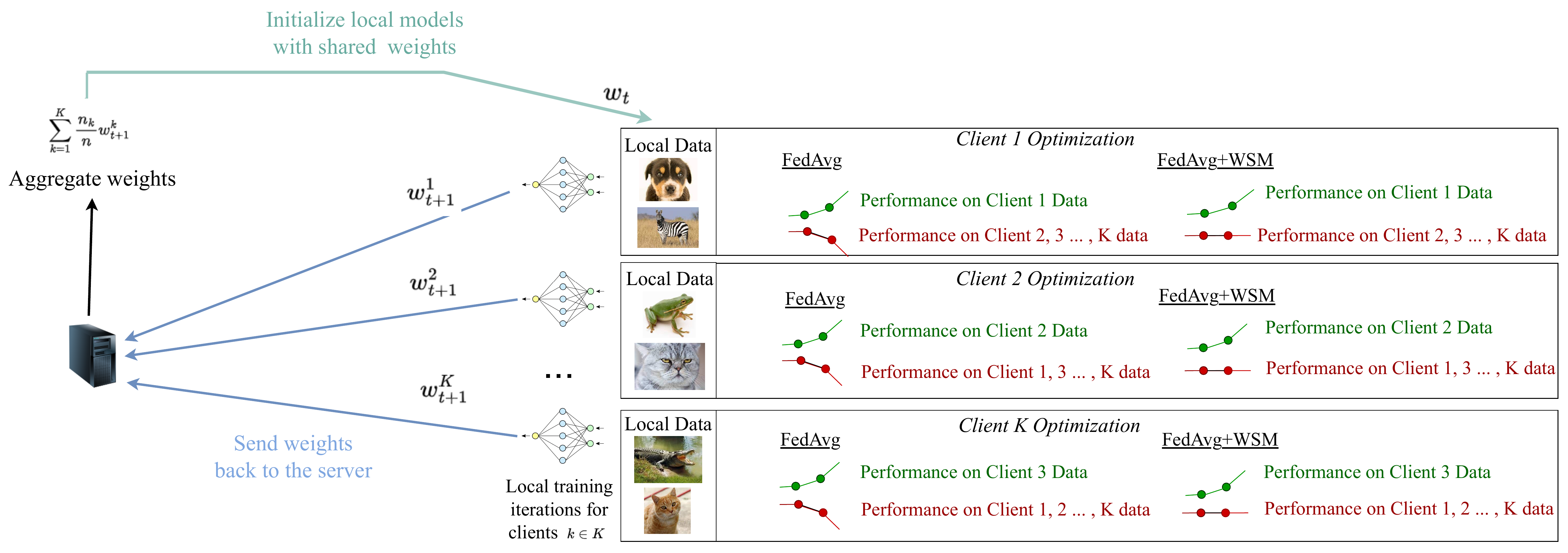}\vspace{-5pt}
    \caption{ \textit{Illustration of catastrophic forgetting within client rounds.} A global model with knowledge of all classes is sent to all clients participating in a given FL round. Local training increases the client model performance on the client's local distribution but tends to simultaneously decrease performance with respect other clients distributions which leads to poor aggregation and overall model performance. \vspace{-12pt}}
    \label{fig:maindiagram}
\end{figure}

In continual learning, a model is trained on a number of tasks sequentially and the learner needs to learn each new task without forgetting knowledge obtained from the preceding tasks. The tendency of a continual learning model to forget previously learned information when learning a new task is termed catastrophic forgetting \citep{mccloskey1989catastrophic} and it is typically a focus of study in continual learning literature. Similar to FL, data heterogeneity in continual learning presents a challenge since different tasks typically contain data drawn from different underlying distributions. 
We can draw a connection between the catastrophic forgetting problem in continual learning and the client drift problem in federated learning. 
We consider one round of federated learning in which $K$ random clients are selected and initialized with a copy of the current global model. Each client performs a pre-determined number of local update steps to optimize the objective on their local data. A round ends with an update to the global model achieved by aggregating the updates from each client. At the beginning of a round, the clients receive a model previously derived from training on other clients data. As local training proceeds, the model becomes increasingly biased towards a given client. As discussed in \citet{lesort2022continual, gupta2022fl}, this can cause client models to rely on spurious correlations to improve their in-distribution performance, creating a situation where local models experience catastrophic forgetting with respect to data of the other clients (drawn from distinctly different distributions). Naturally, aggregating models that have deviated from a joint solution will lead to degraded results with respect to the global objective. We denote this problem as \textit{local client forgetting}. Reducing local client forgetting would moderate the decrease in performance with respect to other clients data at individual client models. We hypothesize this would increase the ability of local models to generalize to the distribution of data from other clients and improve the loss of individual models over the combined data and we therefore propose to reduce client drift by tackling local client forgetting. Figure \ref{fig:maindiagram} illustrates local client forgetting within a round of federated learning. 

There are numerous approaches to tackle catastrophic forgetting in the continual learning literature \citep{kirkpatrick2017overcoming,li2017learning,chaudhry2019tiny,schwarz2018progress,davari2022probing}. Many of these however are largely impractical in the FL setting. Experience Replay methods \citep{chaudhry2019tiny} require access to other clients' data, violating the fundamental data communication constraints of FL. Similar concerns exist for many regularization methods such as elastic weight consolidation (EWC) \citet{kirkpatrick2017overcoming} which require communicating additional information. Regularization methods can additionally require many steps to converge due to the additional conflicting objectives \citep{NEURIPS2019_15825aee}. This computational constraint can hurt convergence of the FL algorithm, a key desideratum. For the supervised continual learning setting, \citet{caccia2022new, ahn2021ss} proposed a modification of the standard cross entropy objective function that truncates the softmax denominator, removing terms corresponding to classes from old tasks. A variant of this method inspired by the long-tailed recognition methods \citep{ren2020balanced} was also recently introduced in \citet{jodelet2022balanced}. This simple approach mitigates catastrophic forgetting by reducing the bias on the model to avoid predicting old classes. 

Inspired by the parallels between client drift in FL and catastrophic forgetting in the continual learning case, we propose an adaptation of the CL method from \citet{caccia2022new, ahn2021ss, jodelet2022balanced} to modify the loss function of each client based on its class distribution using a re-weighted softmax. We will empirically demonstrate that this approach can drastically reduce client level forgetting in the heterogeneous setting, leading to substantially improved overall global model convergence and final performance. We apply the re-weighted softmax to baseline methods FedAvg \citep{mcmahan2017communication}, SCAFFOLD \citep{karimireddy2020scaffold}, FedNova \citep{wang2020tackling} and FedProx \citep{li2020federated} demonstrating performance improvements in all cases. 

\section{Related Work}
\paragraph{Federated Learning} The most commonly used baseline in federated learning is the FedAvg algorithm proposed by \citet{mcmahan2017communication}, which improves on local SGD \citep{stich2018local} by allowing clients to train multiple local iterations successively, thus reducing communication costs. Since the communication cost between two nodes is orders of magnitude larger than the cost between processor and memory on the same node, communication efficiency in federated learning of upmost importance \citep{konevcny2016federated}. Convergence of FedAvg has been widely studied for both i.i.d. \citep{stich2018local, wang2018cooperative} and non i.i.d. settings \citep{karimireddy2020scaffold, li2020federated, fallah2020personalized, yu2019parallel} and settings under which client data is heterogeneous offer additional challenges since convergence has been shown to deteriorate as a function of increasing heterogeneity \citep{li2020federated, hsu2019measuring}.

\paragraph{Non-Heterogeneous Data Partitions and Client Drift}
One significant challenge encountered when training on decentralized data is heterogeneity of samples across clients \citep{konevcny2016federated, li2021survey}. Partitions contain data generated under different conditions, which can reasonably be expected to create different local distributions at each client. For the case of supervised multi-class classification, which is the focus of this work, users may frequently be missing data from an entire class or multiple classes of the global underlying distribution. During FL training, data at each client are sampled
from these local distributions, creating different local objectives. When clients progress too far towards minimizing their own objective, local models drift from one another, degrading the performance of the shared global model and slowing down convergence \citep{yao2021local,li2019convergence,karimireddy2020scaffold,diao2020heterofl}. 

Several attempts have been made to alleviate client drift through various methods. One approach centers around knowledge distillation to regulate local training, \citep{zhu2021data} and \citep{lin2020ensemble} ensemble information about the global data distribution and disseminate it to clients via additional models trained at the server. These methods possess the added risk of privacy attacks and while \citet{zhu2021data} take steps to mitigate this risk, their method requires the existence of an unlabeled dataset, which may not be available in all settings. Other approaches attempt to constrain gradient updates from the clients to reduce the impact of client drift. \citet{karimireddy2020scaffold} propose SCAFFOLD, an algorithm to control client drift using control norms to modify client gradients. The control norms estimate the drift at each client and use that estimate to correct the local updates. FedProx \citep{li2020federated} adds a proximal term to the local objective to limit the impact of variation in local updates. This term is weighted with an additional hyperparameter, $\mu$ which must be tuned appropriately. Both SCAFFOLD and FedProx require careful hyperparameter tuning to be effective \citep{li2022federated}. Like SCAFFOLD and FedProx \citet{acar2021federated} also estimate client drift and use it as a means to constrain updates. However, their operations take place at the server level and corrections are applied to the server updates thus avoiding SCAFFOLD’s use control norms and saving on the inherent communication burden. The general strategy for each of these methods is similar. They attempt to estimate client drift using gradient updates and then constrain the updates to reduce the drift. SCAFFOLD has additionally been shown to have unstable accuracy and has twice the communication burden of FedAvg due to the control norms required for each client \citet{li2022federated}. Our proposed method modifies the objective functions locally, creating no additional communication burdens and introducing no additional hyperparameters. It also directly tackles the problem of the underlying distribution shift, unlike the other methods mentioned which attempt to address client heterogeneity through constrained gradient optimization and/or knowledge distillation. \citet{tenison2022gradient} propose a gradient masking technique that modifies the aggregation of updates on the server side. This method does directly tackle the problem of distribution shift and while it has been shown to be effective at achieving better generalization on non i.i.d. data, our re-weighted softmax is able to take advantage of the structure of the commonly used cross-entropy loss, making it simpler to implement. Additionally, since we modify the objective functions locally, the re-weighted softmax method is compatible with any federated optimization method in the literature. 
% Nice positioning

\vspace{-8pt}\paragraph{Continual Learning}
Continual learning is a process by which tasks are learned sequentially over a period of time and knowledge of previous tasks is retained and leveraged to learn new tasks \citep{chen2018lifelong}. Continual learning is made difficult by the fact that neural networks suffer from catastrophic forgetting, in which learning a new task overrides weights learned from past training, thus degrading model performance on previously learned tasks \citep{mccloskey1989catastrophic}. Several families of methods have been developed to mitigate catastrophic forgetting. The first class of methods are architecture based approaches \citet{schwarz2018progress} that attempt to grow or modify an architecture over time to expand its knowledge. In the second class of methods approaches which store some subset of old data for rehearsal are applied \citep{lopez2017gradient, chaudhry2019tiny,Rebuffi_2017_CVPR}. Finally, a third class of methods regularizes the learned parameters to limit drastic weight changes when learning a new task \citep{kirkpatrick2017overcoming, zenke2017continual}. None of these solutions are widely applicable in a FL setting since they typically require some sort of information sharing across client nodes which is prohibited in the FL framework. A federated continual learning setting has been considered in the literature \citep{yoon2021federated}. Here each client in the federated network continuously collects data. Our work on the other hand considers the standard FL setting where each client maintains a fixed set of data and draws connections to a notion of forgetting across clients to motivate a modification of the loss function. \citet{shoham2019overcoming} have used ideas from continual learning to propose FedCurv, based on the EWC algorithm \citep{kirkpatrick2017overcoming} from continual learning. FedCurv requires sending additional information and is not compatible with all FL methods. Along this line, \citet{xu2022acceleration} also proposed an approach inspired from rehearsal methods, generating pseudo data and adding an additional regularization term. This requires an expensive pseudo data generating procedure and increases the local training time. 

\section{ Methods}

\paragraph{Background}
%Federated optimization refers to the optimization problem implicit to federated learning \cite{konevcny2016federated}. 
In federated optimization, training data is distributed and optimization occurs over $K$ clients with each client $k\in{1, ..., K}$ having data $\mathbf{X}_{k}$ drawn from distribution $D_k$. We define $n_{k}=|\mathbf{X}_{k}|$ and $n=\sum_{k=1}^{K}n_{k}$ for $n$ samples. The data $\mathbf{X}_{k}$ at each node may be drawn from different distributions and/or may be unbalanced with some clients possessing more training samples than others. The typical objective function for federated optimization is given by
\begin{equation} \label{eq:fed_opt}
    \min_{\mathbf{w}\in\mathbb{R}^d}\sum_{k=1}^{K}\frac{n_{k}}{n}\mathcal{L}(\mathbf{w}, \mathbf{X}_{k} ),  %=\mathbb{E}_{k}[F_{k}(w)]
\end{equation}
with $\mathcal{L}(\mathbf{w},\mathbf{X}_{k})$ measuring client $k$'s local objective, and $\mathbf{w}$ representing the global parameters.

In this work we will restrict ourselves to the common case where $\mathcal{L}$ is the cross entropy loss.
There are many possible variations of FL algorithms. In general, they follow the similar structure to FedAvg \citep{mcmahan2017communication}, which proceeds as follows :

\begin{itemize}% [leftmargin=20pt]
    \item \textbf{Client selection:} for a set of $K$ clients, $K*p$ are selected at each round $\{t_{i}\}_{i=1}^{T}$ , where $0 < p \leq 1$ is a pre-determined proportion of clients.
    \item \textbf{Client updates:} At the beginning of round, client models are initialized with the current weights of the server model. Each client selected for the round performs $E$ local iterations of SGD.
    \item \textbf{Cerver update:} The weights of the individual client models are aggregated to form an update to the shared global model.
\end{itemize} 

\paragraph{Re-weighted Softmax Cross Entropy} Consider a neural network $f: \mathcal{R}^D \to \mathcal{R}^C$ where $C$ is the total number of classes.
The standard cross entropy is given by equation \ref{eq:cross_ent} where $y(\mathbf{x})$ is the label of $\mathbf{x}$ and $\mathcal{C}$ is the set of all classes available to the clients.

\begin{align} \label{eq:cross_ent}
    \mathcal{L}_{CE}(\mathbf{X}_k,\mathbf{w})&=-\sum_{\mathbf{x} \in \mathbf{X}} \log\frac{\exp(f_{\mathbf{w}}(\mathbf{x})_{y(\mathbf{x})})}{\sum_{c\in \mathcal{C}} \exp(f_{\mathbf{w}}(\mathbf{x})_{c})} \\
    &=-\sum_{\mathbf{x} \in \mathbf{X}} \Bigl[f_{\mathbf{w}}(\mathbf{x})_{y(\mathbf{x})} - \log\Bigl(\sum_{c\in \mathcal{C}} \exp(f_{\mathbf{w}}(\mathbf{x})_{c}\Bigr)\Bigr]
\end{align}

One interpretation of this classical loss function considers the two terms as a tightness term (the first term) which brings samples close to their representative classes  and a contrast term (the second term) which pushes them apart from other classes \citep{boudiaf2020unifying}. We note similar loss functions can be interpreted from an energy modeling view \citep{liu2020energy}. 
%% Add energy loss function and gradient stuff here?? 

We now modify the standard cross entropy using the re-weighted softmax (WSM) to give our per-client objective function in Equation \ref{eq:wsm} where $\boldsymbol{\mathbf{\beta}_k}$ is a vector containing the proportions of each class present in the client dataset and $\mathbf{\beta}_c\in \boldsymbol{\beta_k}$ is the proportion of label $c$ present in the dataset. %While this reformulation makes WSM non probabilistic, a probabilistic interpretation is not a strict requirement of loss function.  Additionally, 

%We can consider an equivalent energy function formulation of this loss for WSM that we include in \ref{sec:energy_loss}.

\begin{equation}\label{eq:wsm} 
\mathcal{L}_{WSM}(\mathbf{X}_k, \mathbf{w}) =  
-\sum_{\mathbf{x} \in \mathbf{X}}\Bigl[ f_{\mathbf{w}}(\mathbf{x})_{y(\mathbf{x})} - \log\Bigl(\sum_{c\in \mathcal{C}} \beta_c\exp(f_{\mathbf{w}}(\mathbf{x})_{c})\Bigr)\Bigr]\end{equation}

We note that in Equation \ref{eq:wsm} the weighting $\beta$ introduces in the second term is a function only of the labels present in the client data $\mathbf{Y}_k$. In a highly imbalanced class scenario, as often studied for FL, many $\beta_c$ will be zero or contain very small values, thus removing or substantially degrading the contribution of that class to the contrast term. Intuitively, aggressively optimizing $\mathcal{L}_{CE}$ through multiple gradient steps during a client round can lead to a drastic increase  in $\mathbb{E}_{x,y\sim D_{j \neq k}}[l_{CE}(\mathbf{x},\mathbf{y})]$ where $D_j$ are the distribution of clients other than client $k$. This is because the contrast term encourages classes not present on the client to never be predicted. The re-weighted softmax approach modifies the original local objective function to avoid excessive pressure that drives up the loss of other client data.  Indeed, classes not present at the current client are ignored by the local optimization which forces the client to learn by adapting the model's internal representation of the classes present in its training data, rather than abruptly shifting representations of classes outside its training set \citep{caccia2022new}. We will demonstrate empirically this leads to a reduction in local client forgetting defined below.

\paragraph{Local client forgetting} We formalize the notion of local client forgetting discussed in Sec~\ref{sec:intro} for a multi-class classification problem. Denoting the accuracy on a client $k$'s local test data $Acc_k(\mathbf{w})$, where $\mathbf{w}$ are the model parameters.  We can define local client forgetting according to equation \ref{eq:forget} where $\mathbf{w}_t^i$ refers to the model of client $i$ at round $t$ after it has completed local training (prior to aggregation) and $\mathbf{w}_{t-1}$ is the global model (after aggregation) at the end of round $t-1$.
\begin{equation} \label{eq:forget}
    F_{ki}=Acc_k(\mathbf{w}_{t-1}) - Acc_k(\mathbf{w}_{t}^i)
\end{equation}. 
 
\noindent We define an average forgetting for a client $k$'s model according to equation \ref{eq:avg_forget}
\begin{equation}\label{eq:avg_forget}
    F_k = \frac{1}{K-1} \sum_{i \neq k} F_{ki}
\end{equation}
\noindent In what follows we will study these quantities for a standard FL setting.

\section{Experiments}\label{sec:exp}
In this section, we present the empirical results for the WSM framework. We start by analyzing the notion of forgetting in the context of standard FedAvg, and then show how the application of the re-weighted softmax objective can resolve this. Subsequently, we study how WSM can enhance standard FL algorithms, improving their overall performance. 

\begin{wrapfigure}{r}{0.36\textwidth}
    \centering
    \includegraphics[width=0.3\textwidth]{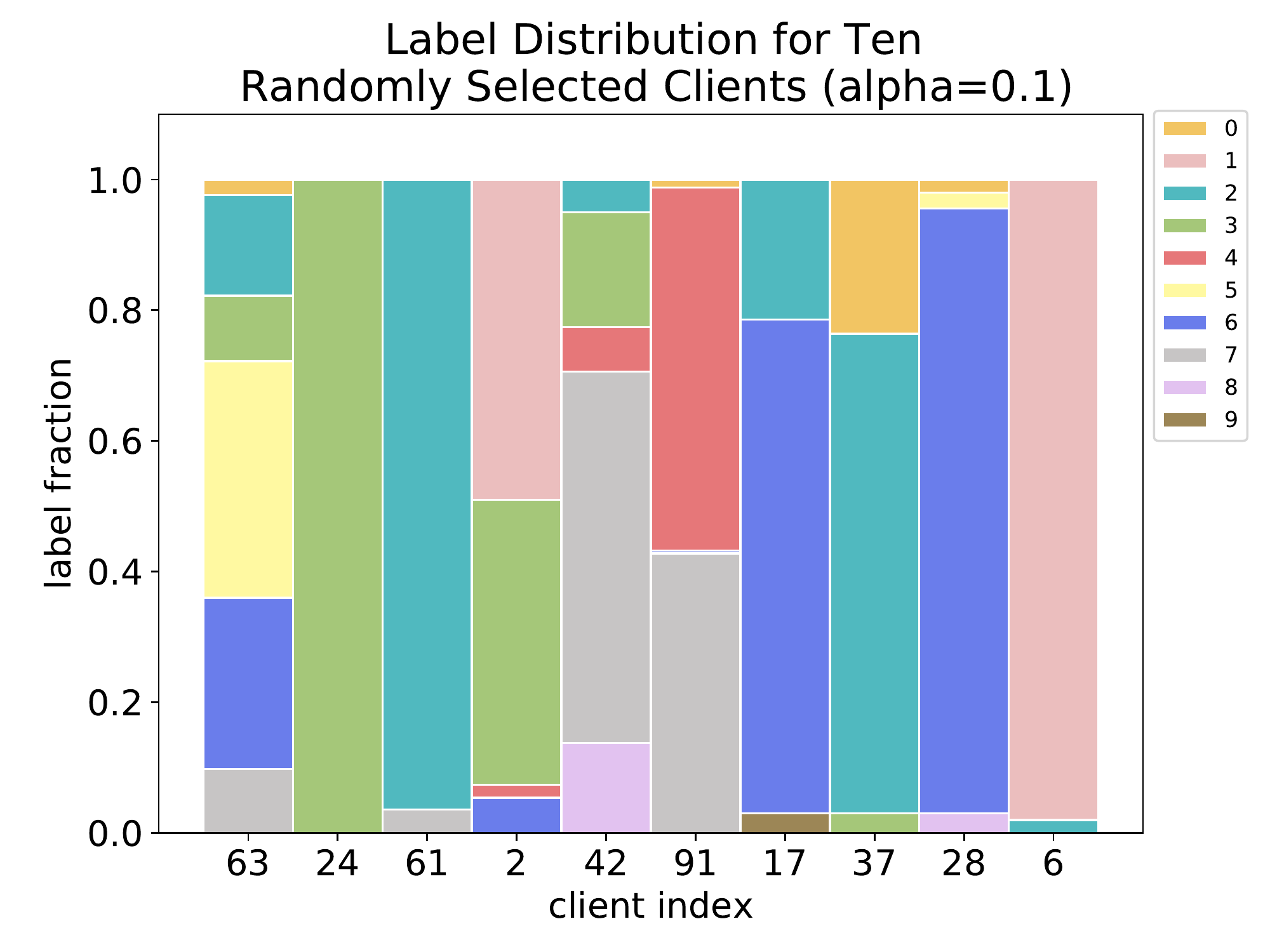}
    \caption{CIFAR-10  - Distributions of ten randomly selected clients with data partitioned according to a Dirichlet distribution parameterized by $\alpha=0.1$. \vspace{-1cm}}
    \label{fig:alpha01}
\end{wrapfigure}

\paragraph{Datasets and Data Partitions} 
We utilize CIFAR-10, CIFAR-100 \citep{Krizhevsky_2009_17719} and FEMNIST \citep{caldas2018leaf} datasets in our experiments, each of these datasets come pre-separated into training and testing sets. Our primary evaluations consider 100 clients where each client requires their own training and validation sets according to their own unique distribution. To facilitate this, the entire training set is separated into equally sized non-i.i.d. partitions using the Dirichlet distribution parameterized by $\alpha=0.1$, similar to the method of \citet{hsu2019measuring}. To provide intuition into what client partitions parameterized by $\alpha=0.1$ look like in practice, Fig. \ref{fig:alpha01} shows the data distributions of ten randomly selected clients. These client partitions are then further separated into training ($90\%$) and validation ($10\%$) sets for each client. For example, 100 clients being trained using CIFAR-10 which contains 50 000 training samples would each have 500 of these training samples. Of those 500 samples, 450 would be used for local model updates and 50 would be used exclusively for validation. 
\paragraph{Settings}
We follow the experimental settings of \citet{reddi2020adaptive}. Clients are sampled without replacement for each round but can be selected again in subsequent rounds. The fraction of clients sampled is $10\%$ for CIFAR-10 and FEMNIST datasets and $2\%$ for CIFAR-100. Our primary evaluations train a ResNet-18 over 4000 communication rounds for 3 local epochs, using a mini-batch of size 64 and a learning rate of 0.05 for CIFAR-10 and CIFAR-100. FEMNIST, which converges faster due to its large size, is trained for 3000 rounds with all other settings the same as for the CIFAR datasets. We use SGD as our optimizer, with weight decay of $1\times10^{-4}$ following \citet{yao2021local, hsu2019measuring}. In these experiments, we observe that FedAvg often (though not always) performs better with group normalization as indicated by \citet{hsieh2020non} while FedAvg+WSM is able to perform well with both group and batch normalization, very frequently achieving the best results with batchnormalization. We therefore treat the normalization method as a hyperparameter and provide the best results for both batch and group normalization for both methods at each learning rate.

\vspace{-10pt}\paragraph{Validation}
Throughout the training process the global model is periodically evaluated on the aggregation of the client validation sets to gauge overall training progress, the test set on the other hand is only used at the end of the training process. For lower values of $\alpha$, as client distributions become more skewed, there can be significant changes in accuracy between training runs \citep{hsu2019measuring}. 
Since we focus our analysis on the highly heterogeneous case in which the Dirichlet distributions at each client are parameterized by $\alpha=0.1$, we observe higher variance in our results. This is especially true for smaller datasets such as CIFAR-10 and CIFAR-100. To mitigate these effects on the validation statistics, we follow the lead of \citet{reddi2020adaptive} and report our final accuracy as the average of the test accuracies taken over the last 100 rounds of training.

\begin{figure}[t]
     \centering 
    % \begin{subfigure}[b]{0.35\textwidth}
     %    \centering
      %   \includegraphics[width=1.1\textwidth]{forget.pdf}
      %   \vspace{1.5cm}
   %  \end{subfigure}
    % \begin{subfigure}[b]{0.55\textwidth}
        % \centering
         \includegraphics[width=.32\textwidth]{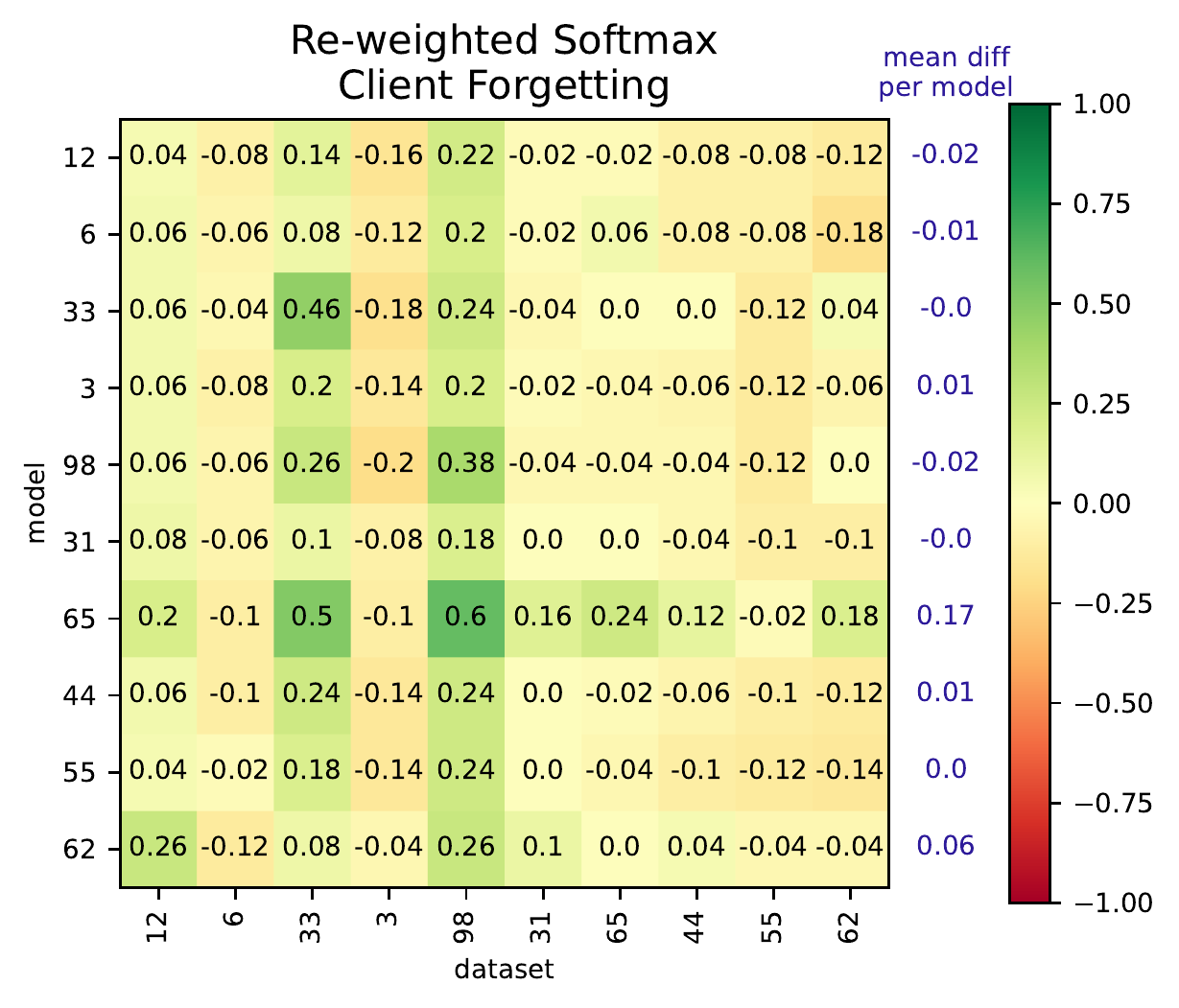}%round 1200 diff
        \includegraphics[width=.32\textwidth]{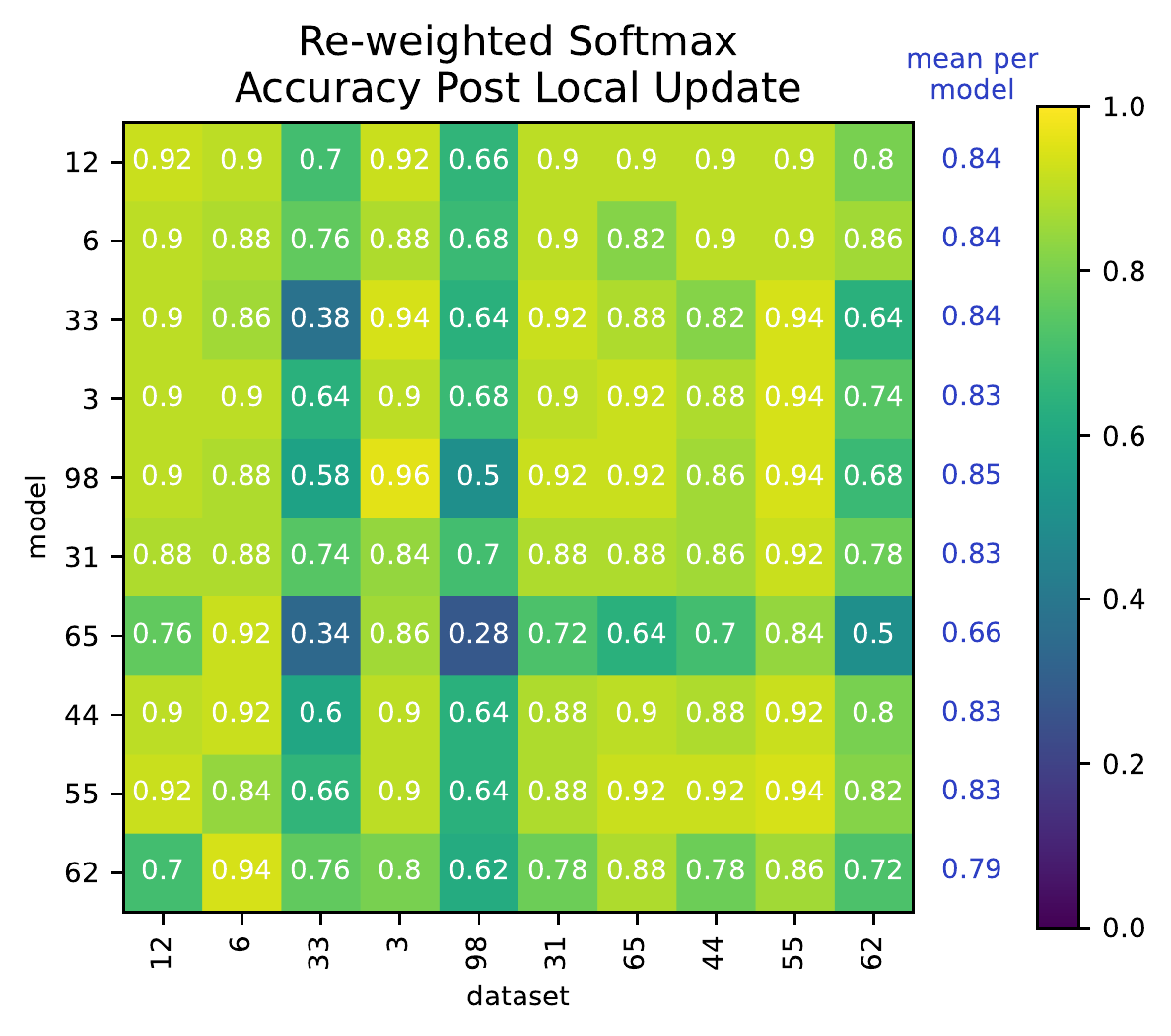}\\%round 1200 post
        \includegraphics[width=.32\textwidth]{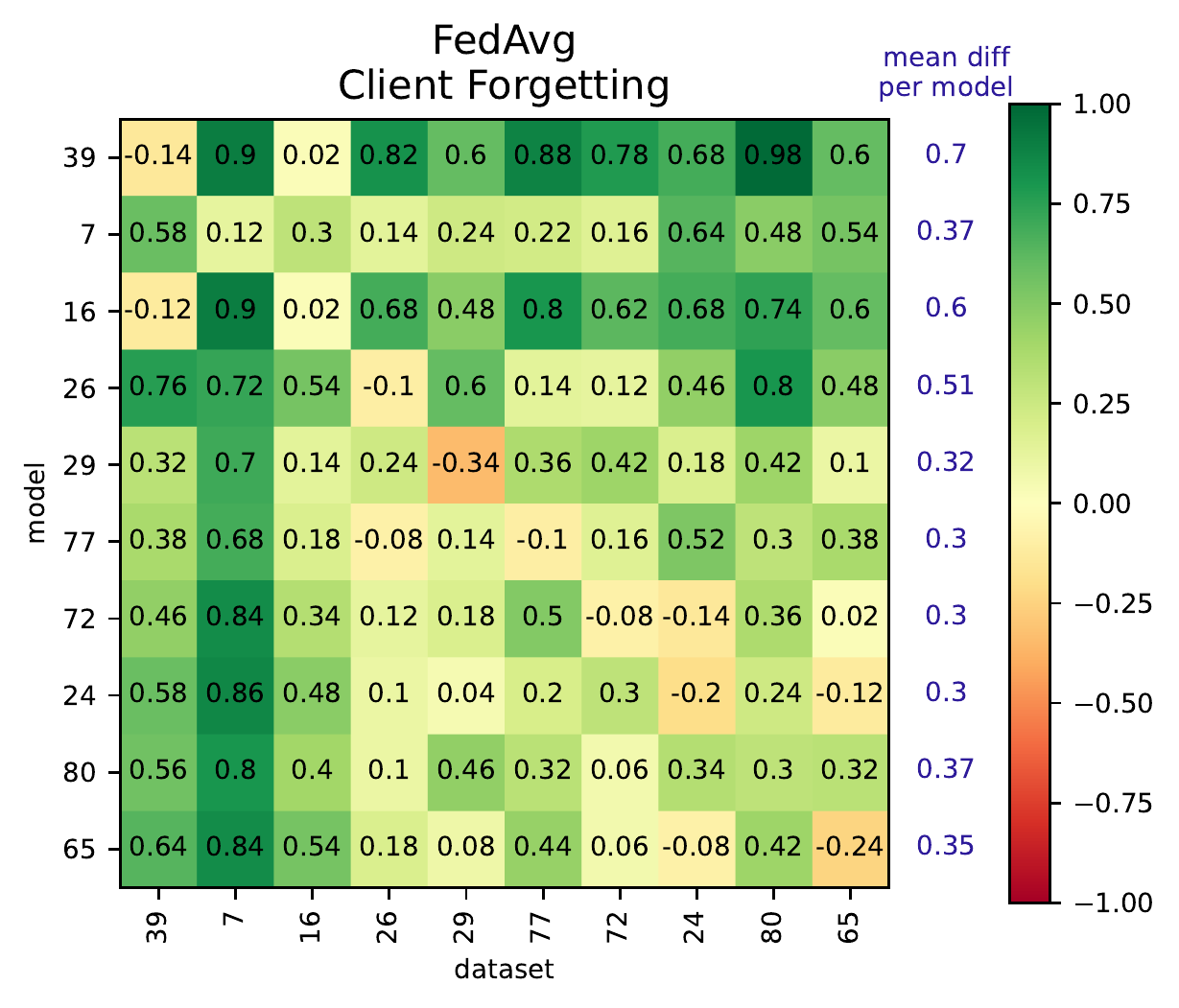}%round 1200 diff
        \includegraphics[width=.32\textwidth]{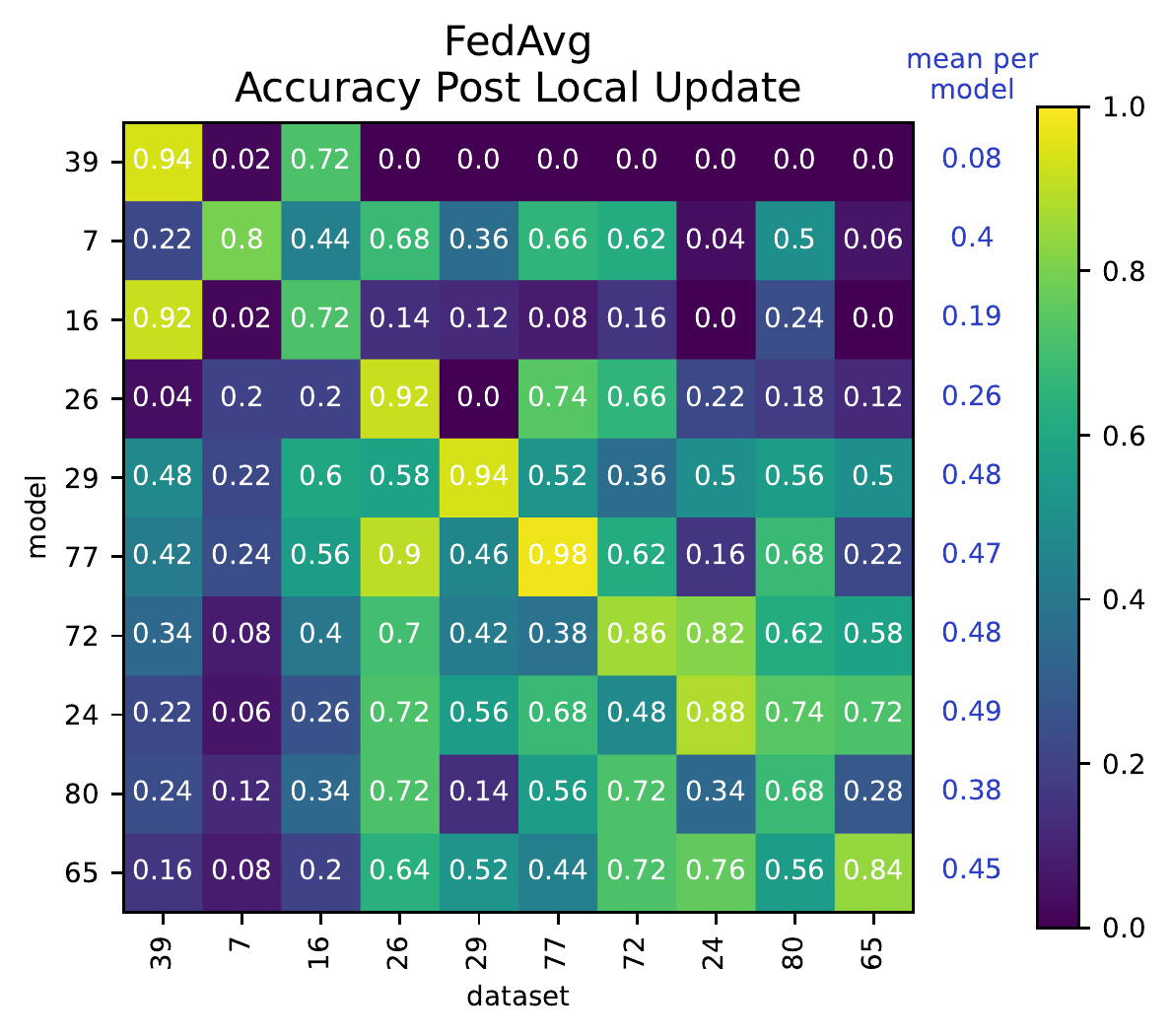}%round 1200 post
        \vspace{-12pt}
    % \end{subfigure}
     \caption{\small \textit{Illustrating local client forgetting for a given round with and without WSM}. For each heatmap the y-axis indicates the model of client i and the x-axis indicates the local data of client k. In the left column we show the $F_{ik}$ as defined in Eq. \ref{eq:forget}. Positive values of forgetting (green) indicate high forgetting.  The right column show the accuracy of client i's model after training when evaluated on client k's data. We note that in some cases the accuracy can completely collapse on other client's data (particularly when they don't overlap in any classes). We see that FedAvg+WSM (top row) significantly reduces forgetting across clients.\vspace{-0.72cm}}
     \label{fig:client_forget}
\end{figure}  

\vspace{-12pt}\subsection{Forgetting During a Federated Round}\vspace{-5pt}
We first study and build intuition on the effects of forgetting in a federated round for CIFAR-10 dataset. In Figure~\ref{fig:client_forget} we show the effects of forgetting during a round of federated learning. The heatmaps on the bottom row of Figure~\ref{fig:client_forget} are for round 1200 of training. In the forgetting heatmap we observe a lot of green in the off-diagonal terms indicating that forgetting is very high when using standard cross entropy. 

The post local update heatmap for standard cross entropy shows a strong trend of better accuracies along the diagonal indicating model $i$ does much better on dataset $k$ when $i=k$, its own dataset.
\begin{wrapfigure}{r}{0.4\textwidth}
    \includegraphics[width=0.4\textwidth]{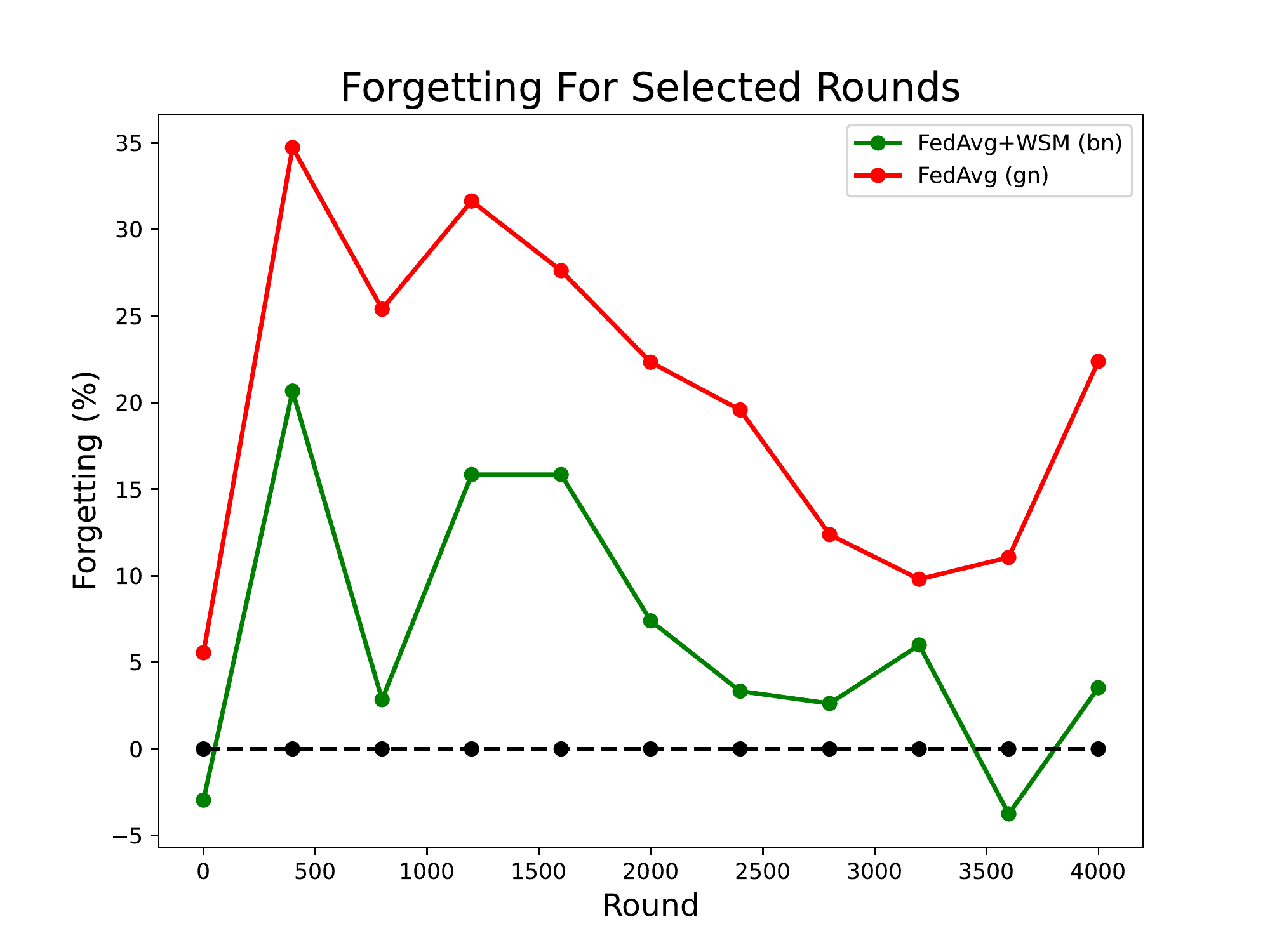}\vspace{-5pt}
    \caption{\small Average forgetting $F_{k}$  for selected rounds of training. We observe high forgetting for FedAvg which is substnatially reduced (especially towards the end of training when applying WSM.\vspace{-0.5cm}}
    \label{fig:forgetPlot}
\end{wrapfigure}
 When using WSM, we observe no preference for better accuracy along the diagonal as is the case without using WSM. In fact, we notice lower accuracies to be concentrated along the column, indicating a particularly difficult dataset for all local models or along the row, indicating a model that does badly on all datasets including its own. These observations are supported by the forgetting heatmap for WSM which is predominantly yellow indicating very low forgetting values. 
These results show WSM has the ability to greatly limit the effects of local client forgetting by ignoring classes outside of its data distribution and focusing learning on the classes present. This narrower focus leads to better overall performance after aggregation. The heatmaps are shown for round 1200 of training, approximately a third of the way through training, however, the observation is further confirmed for other rounds as shown in Figure~\ref{fig:forgetPlot} where we plot forgetting throughout training. We observe on average it is high in FedAVG and is substantially reduced especially at the end of training for WSM. Having shown that WSM can indeed reduce the local client forgetting, we now study its effect on the aggregated models. %The data shown in figure~\ref{fig:client_forget} is for round 1200 of training but these features are evident throughout the training cycle which we show by providing additional heatmaps in the appendix (\ref{sec:additional_forgetting}). 
\vspace{-4pt}
\subsection{Evaluation of WSM}\vspace{-4pt}
In this section we demonstrate how WSM used in combination with FedAvg can improve model performance and convergence. Table~\ref{tab:main} shows model performance across a range of learning rates, using CIFAR-10, CIFAR-100, and FEMNIST datasets. 
%For this evaluation we treat the normalization method as a hyperparameter and report the accuracy using the best performing normalization method which is batch norm for WSM and group norm for FedAvg. Section~\ref{sec:main_eval_plus} contains additional results including the evaluation of FedAvg and FedAvg+WSM for both batch norm and group norm. % EB: we dont need to directly reference this again as its already mentioned higher up, if they want to look for these details they can find in appendix
The reported values are the average $\pm$ the standard deviation of three runs for each setting. The best preforming model for each learning rate is shown in bold and the best overall result for each dataset is indicated by a green (red) box for WSM (FedAvg). This table demonstrates that WSM has a few benefits to offer during training. WSM substantially improves performance for both CIFAR datasets with a $2.2\%$ and $1.3\%$ improvement for CIFAR-10 and CIFAR-100, respectively. For FEMNIST dataset the results show the best performing models are within statistical error of one another. We do however observe strong results using WSM with higher learning rates under which regime we are able to obtain faster convergence.  WSM also makes the hyperparameters easier to tune since it performs well over a large range of learning rates, for example for learning rates between $0.03$ and $0.1$ FedAvg+WSM has remarkably steady performance between $85.5\%$ to $85.7\%$ while we observe no such consistent performance for vanilla FedAvg.

\begin{table*}[t]
\caption{\small Accuracy results of FedAvg with and without WSM for different hyperparameters. We observe that FedAvg+WSM consistently improves performance over FedAvg over a wider range, as well as having the highest overall accuracy by a substantial margin on the cifar datasets. WSM also makes the learning rate easier to tune since we observe a large hyperparameter range \vspace{-5pt}}
\scriptsize
\begin{center}
\begin{tabular}{ ccccc } 

\toprule
\multirow{ 2}{*}{Method}  &  & \multicolumn{3}{c}{\emph{\ \ \ \ \ \ \ Dataset}} \\
 &lr &CIFAR-10 &CIFAR-100 &FEMNIST \\ 
 \midrule
\textsc{ FedAvg } & 0.5 &$0.326\pm 0.098$  & $0.292\pm 0.012$ & $0.542\pm 0.087$ \\
\textsc{ FedAvg+WSM (ours)}&  &$\mathbf{0.792\pm 0.006}$& $\mathbf{0.426\pm 0.003}$& $\mathbf{0.837\pm 0.002}$\\
\cdashline{1-5}
\textsc{ FedAvg } & 0.3 & $0.791\pm 0.013$  & $0.384\pm 0.013$ & $0.769\pm 0.006$ \\
\textsc{ FedAvg+WSM (ours)}& &$\mathbf{0.834\pm0.008}$ &$\mathbf{0.467\pm0.015}$& $\mathbf{0.844\pm 0.010}$\\
\cdashline{1-5}
\textsc{ FedAvg } &0.1 &$0.724\pm 0.027$  & \fcolorbox{red}{white}{$0.500\pm 0.016$} & $0.835\pm 0.002$ \\
\textsc{ FedAvg+WSM (ours)}& &$\mathbf{0.855\pm0.004}$ &$\mathbf{0.514\pm0.009}$&\fcolorbox{green}{white}{ $\mathbf{0.848\pm 0.006}$}\\
\cdashline{1-5}
\textsc{ FedAvg } & 0.07 &$0.826\pm 0.007$  & $0.437\pm 0.007$ &  $\mathbf{0.827\pm 0.006}$  \\
\textsc{ FedAvg+WSM (ours)}& &$\mathbf{0.856\pm0.005}$ &$\mathbf{0.553\pm0.018}$& $0.826\pm 0.019$\\
\cdashline{1-5}
\textsc{ FedAvg } & 0.05& $0.827\pm 0.004$  & $0.464\pm 0.001$ & \fcolorbox{red}{white}{$\mathbf{0.853\pm 0.004}$} \\
\textsc{ FedAvg+WSM (ours)}& &\fcolorbox{green}{white}{$\mathbf{0.858\pm0.003}$} &$\mathbf{0.564\pm0.007}$& $0.842\pm 0.005$\\
\cdashline{1-5}
\textsc{ FedAvg } & 0.03 &\fcolorbox{red}{white}{$0.836\pm 0.005$} & $0.431\pm 0.020$ &  $\mathbf{0.835\pm0.006}$\\
\textsc{ FedAvg+WSM (ours)}& &$\mathbf{0.857\pm0.005}$ &\fcolorbox{green}{white}{$\mathbf{0.581\pm0.005}$}&$0.834\pm 0.003$\\
\cdashline{1-5}
\textsc{ FedAvg } & 0.01 &$0.815\pm 0.003$  & $0.431\pm 0.005$ & $\mathbf{0.830\pm0.002}$ \\
\textsc{ FedAvg+WSM (ours)}& &$\mathbf{0.845\pm0.006}$ &$\mathbf{0.574\pm0.006}$& $0.800\pm0.019$\\
\cdashline{1-5}
\textsc{ FedAvg } & 0.007 &$0.817\pm 0.007$  & $0.426\pm 0.005$ & $\mathbf{0.821\pm 0.011}$ \\
\textsc{ FedAvg+WSM (ours)}& &$\mathbf{0.841\pm0.004}$ &$\mathbf{0.568\pm0.003}$& $0.800\pm0.007$ \\
\cdashline{1-5}
\textsc{ FedAvg } & 0.005 &$0.802\pm 0.010$  & $0.426\pm 0.002$ & $\mathbf{0.819\pm0.022}$ \\
\textsc{ FedAvg+WSM (ours)}& &$\mathbf{0.826\pm0.009}$ &$\mathbf{0.554\pm0.004}$& $0.773\pm0.013$\\
\bottomrule
\end{tabular}
\end{center}
\label{tab:main}
\vspace{-5pt}
\end{table*}

\begin{figure*}[t]
    \centering
    \includegraphics[width=.48\textwidth]{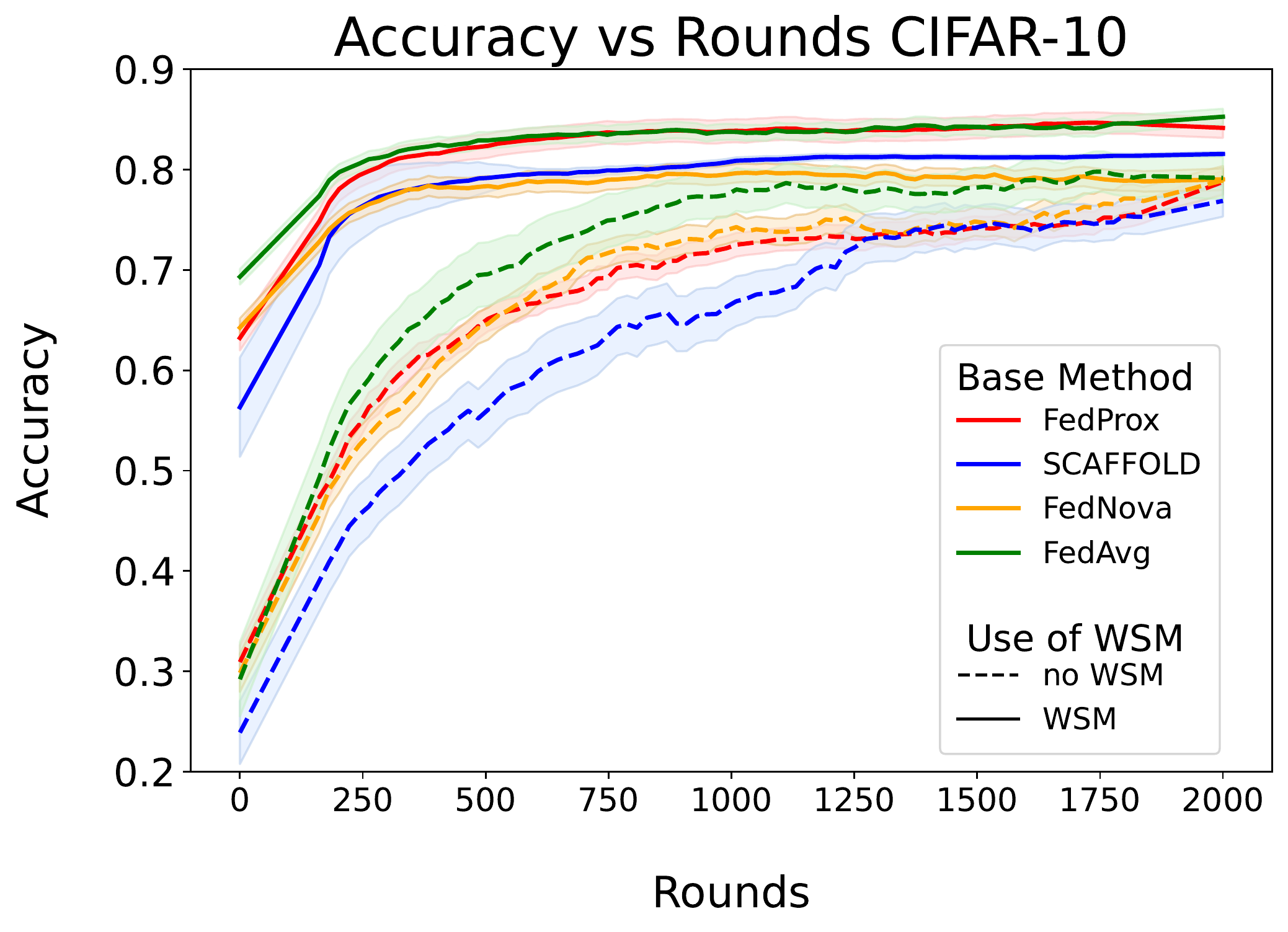}
    \includegraphics[width=.48\textwidth]{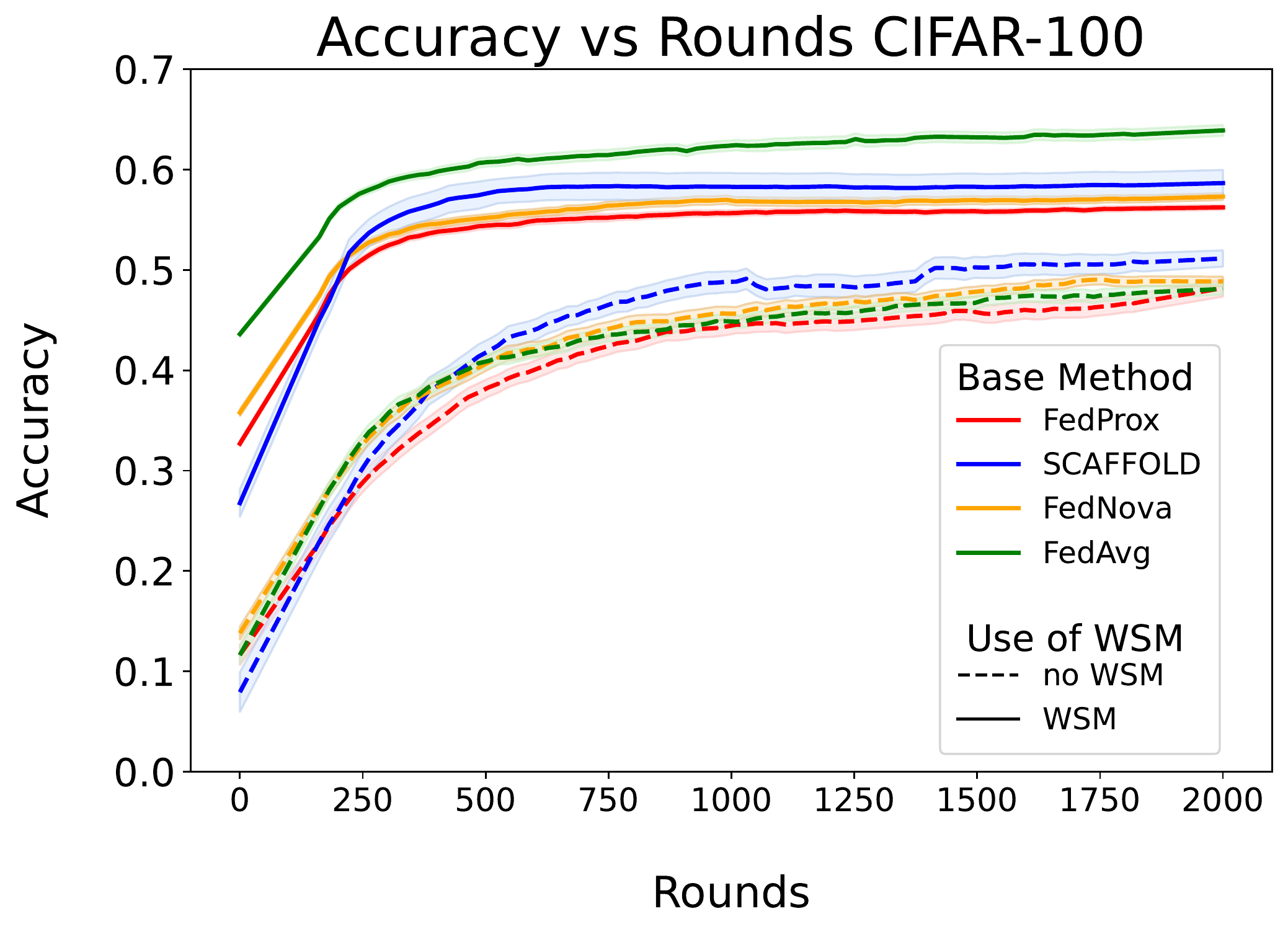}
    \caption{Convergence plots of different algorithms with and without WSM. We observer that WSM variants lead to substantially better convergence for all the compared methods.\vspace{-15pt}} 
    
    \label{fig:main}
\end{figure*}

\begin{table*}[t]

\begin{center}
\caption{\small Accuracy results of FedAvg \citep{mcmahan2017communication}, SCAFFOLD \citep{karimireddy2020scaffold} FedNova \citep{wang2020tackling} and FedProx \citep{li2020federated} with and without WSM. We observe that even combined with FL optimization based methods designed to address heterogeneity, WSM provides consistent performance gains. Furthermore, the best performance overall is highlighted in red. We find that although SCAFFOLD and FedPROX can handle heterogeneity better than FedAvg, when combined with WSM FedAvg can obtain the best performance amongst all methods.} \label{tab:method_compare}
\footnotesize
\begin{tabular}{ccc} 
\toprule
%& & \emph{Dataset} &\\
Method &CIFAR-10 &CIFAR-100 \\ 
 \midrule
\textsc{  FedAvg} &$0.800\pm 0.033$ & $0.494\pm 0.008$ \\
\textsc{ FedAvg+WSM (ours)} &$\mathbf{0.864\pm 0.008}$ &$\mathbf{0.640\pm 0.002}$ \\\cdashline{1-3}
\textsc{  Scaffold} &$0.794\pm 0.033$ &$0.532\pm 0.003$  \\
\textsc{ Scaffold+WSM (ours)} &$\mathbf{0.812\pm0.012}$ &$\mathbf{0.572\pm0.028}$ \\\cdashline{1-3}
\textsc{ FedNova} &$0.780\pm 0.010$ &$0.500\pm 0.005$  \\
\textsc{ FedNova+WSM (ours)} &$\mathbf{0.789\pm0.13}$ &$\mathbf{0.575\pm 0.002}$ \\\cdashline{1-3}
\textsc{  FedProx} &$0.776\pm 0.028$ &$0.487\pm 0.10$  \\
\textsc{ FedProx+WSM (ours)} &$\mathbf{0.818\pm 0.023}$ &$\mathbf{0.562\pm 0.005}$ \\
\bottomrule
\end{tabular}
\end{center}
\vspace{-15pt}
\end{table*}

\vspace{-8pt}\paragraph{WSM for Heterogenous FL methods} So far we have demonstrated the ability of WSM to improve model preformance for the FedAvg algorithm, now we demonstrate its effectiveness over a range of FL algorithms.  In Table~\ref{tab:method_compare} we apply WSM to SCAFFOLD, Fed Nova and FedProx. SCAFFOLD and FedProx are constrained optimization based methods, specifically designed to address the problem of data heterogeneity and Fed Nova as well is designed to improve performance on heterogeneous data. 
%FedNova, SCAFFOLD and FedProx perform better with smaller learning rates so we decrease the learning rate for these algorithms to $0.01$ while we keeping $0.05$ for FedAvg, the additional FedProx hyperparameter $\lambda$ was determined by a hyperparameter search and is set to $1.1$. 
In these experiments we use 30 clients and train the models over only 2000 rounds, since some of these algorithms have a high overhead and many hyperparameters to search. We keep the same local batch size of 64 and 3 local iterations from the base case.  Combining each of the methods with WSM we confirm our hypothesis that using WSM to mitigate local client forgetting provides improvement over the base cases for each method, in most cases the improvement is substantial. From Figure~\ref{fig:main} we remark that combining WSM with a baseline method generally provides a stronger performance from the very beginning of training since for both CIFAR-10 and CIFAR-100 the curves showing the training progression for the methods combined with WSM all start off with higher reported accuracy than the baselines and this improvement in performance persists for the duration of training. Work on critical learning periods, where critical learning periods are defined as the early epochs of a training regime, have shown they can determine the final quality of a deep neural network for traditional ML methods \citep{jastrzkebski2018relation, achille2017critical,yan2021critical} investigate critical learning periods in the FL setting and discover they do indeed exist consistently in FL. We can thus hypothesize that the early training advantage we see when applying WSM may be having a positive impact on its consistent ability to outperform other FL algorithms it is applied to. 

The ease of application of WSM is a significant advantage since this feature makes it an easy way to achieve a substantial improvement over a baseline algorithm. Finally, we observe that combining WSM with FedAvg has the largest overall effect since the best overall method in table~\ref{tab:method_compare} is FedAvg+WSM. While WSM does offer improvements over all baseline algorithms, this result is somewhat surprising since our intuition was that WSM would work independently alleviate local client forgetting and would augment the value of the modifications offered by the different federated learning algorithms. We theorize that the superior performance of FedAvg+WSM over SCAFFOLD+WSM or FedProx+WSM or Fed Nova+WSM is because while SCAFFOLD, Fed Nova and FedProx with WSM all individually improve the baseline, they may need to be further optimized to work effectively together.

\begin{figure*}[t]
    \centering
    %Consider removing
    \includegraphics[width=.32\textwidth]{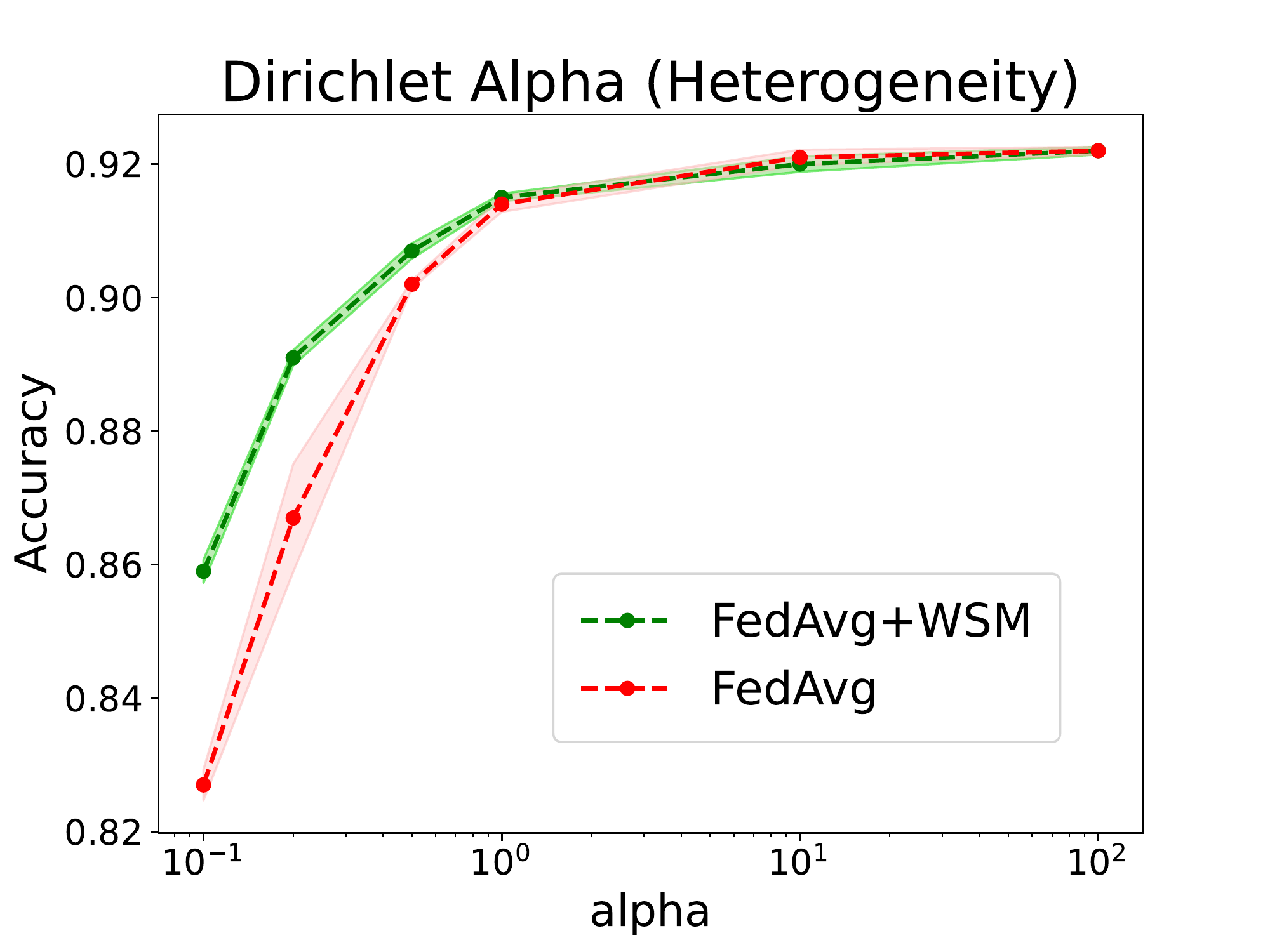}
    \includegraphics[width=.32\textwidth]{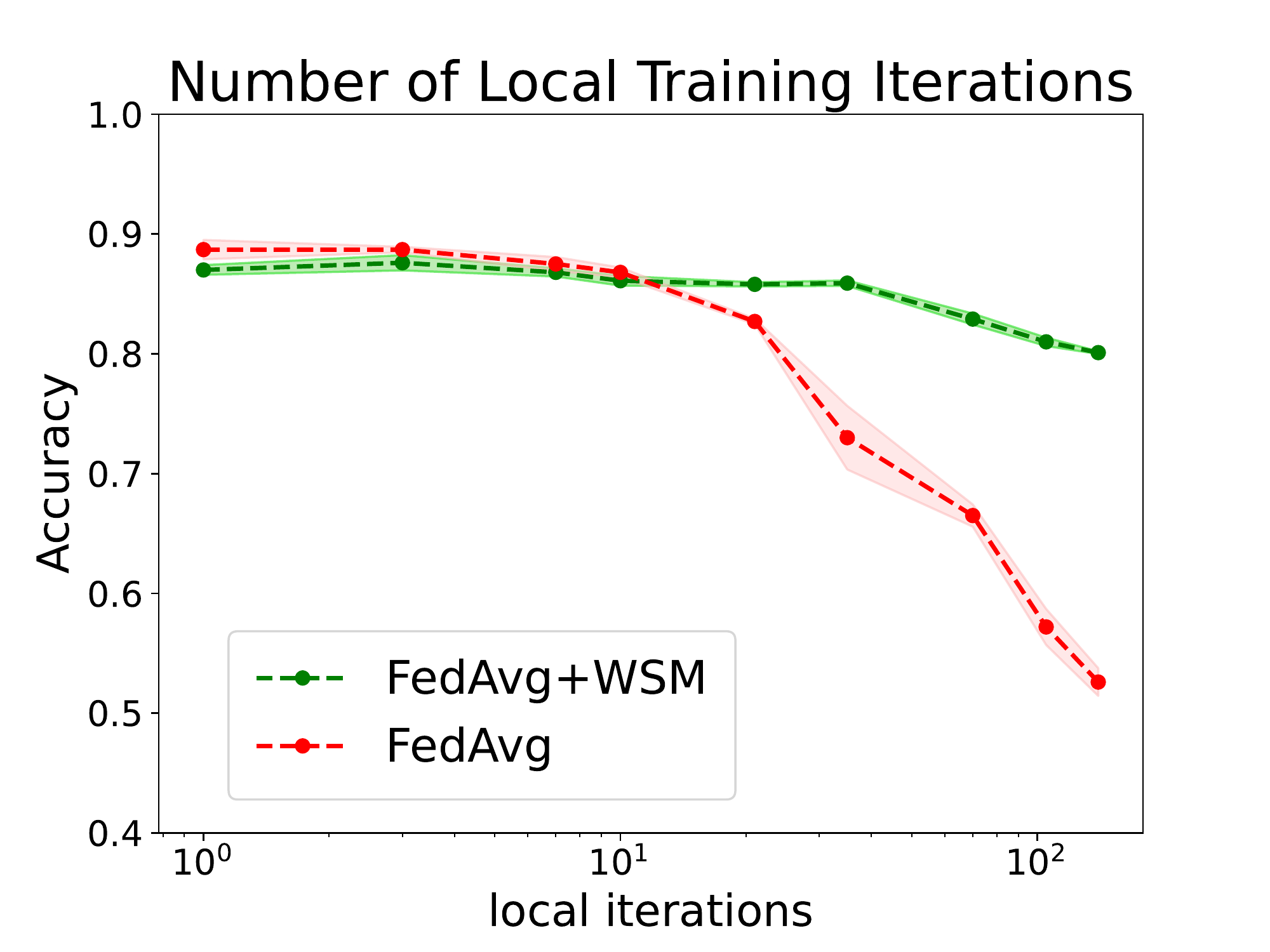}
    \includegraphics[width=.32\textwidth]{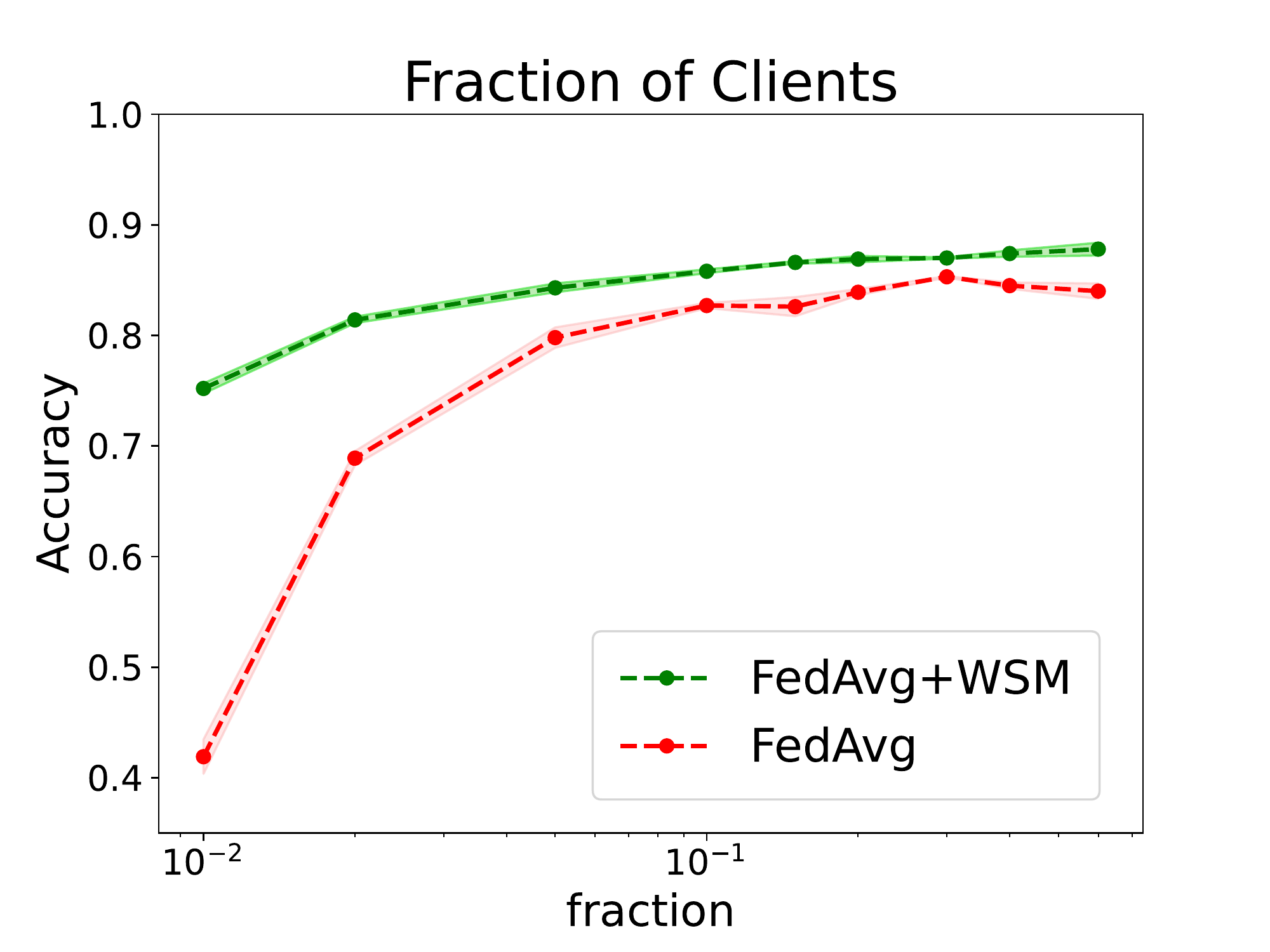}
    \caption{\small Plots of ablations of dataset heterogeneity, number of local iterations and the fraction of clients selected at each round studied with WSM. We observe WSM provides performance increases under most of the conditions studied but it's most significant advantages are in scenarios with very heterogeneous  client data and smaller fractions of clients are selected for each training round.  \vspace{-1cm}} 
    
    \label{fig:ablations}
\end{figure*}
\vspace{-10pt}\subsection{Ablations}\vspace{-5pt}
We now further study the behavior of WSM in combination with FedAvg under different data distributions, client participation settings, different numbers of local iterations and different model architectures. In Figure~\ref{fig:ablations} ablations for the CIFAR-10 setting are shown where we ablate one setting at a time. Except for where we specify the value of the parameter being ablated, the hyperparameters for the ablation studies is the same as for our base case described in section \ref{sec:exp}. We observe that under conditions of low $\alpha$ (high heterogeneity)  and low fractions of clients selected at each round the use of WSM is particularly advantageous.  

\vspace{-8pt}\paragraph{Parameter $\alpha$ of the Dirichlet Distribution} The Dirichlet distribution is parameterized by $\alpha$ as $\alpha\rightarrow0$ the client distributions will become increasingly heterogeneous and as $\alpha\rightarrow\infty$, each client data 
 edges closer towards the same i.i.d. distribution. In our base case we use $\alpha=0.1$ and we now investigate how model performance is affected by changing the heterogeneity of the data partitions. Similar to the work of \citet{hsu2019measuring}, we investigate $\alpha = \{0.01, 0.1, 0.2, 0.5, 1, 10, 100\}$, for a more explicit indication of how client distributions change as a function of $\alpha$ we refer the reader to section~\ref{sec:dir_alphas}.  From Figure \ref{fig:ablations} we observe WSM has a more significant effect as $\alpha$ decreases with the biggest margin of improvement over vanilla FedAvg occurring when $\alpha=0.01$. As the data becomes increasingly homogeneous, the gap between cross entropy with and without WSM shrinks until the data is i.i.d. and the two methods are equivalent. From Figure \ref{fig:alphas} we see that by the time $\alpha=0.5$ most of the clients possess all of the classes, albeit in very skewed proportions. This is also the point at which the performance of FedAvg and FedAvg+WSM become very close. While WSM continues to offer a performance increase over the entire range of $\alpha$ except for $\alpha=100$, we conclude WSM is most advantageous when clients labels distributions are imbalanced.

\vspace{-8pt}\paragraph{Fraction of Participating Clients} One of the characteristics of FL is that data is massively distributed and nodes may have limited communication with the central server as nodes may be offline or have slow connections limiting their communication\citep{mcmahan2017communication}. The direct result of the limited capacity for connection is that only a small number of client nodes may participate in updates at any given time. These ablations focus on the fraction of clients participating in an update round, for each experiment the fraction of participating clients is constant for all rounds of training and we explore fractions from $1\%$, where only one client participates in the update, up to $60\%$. For the fraction of participating clients $p$, we observe the largest performance gap between FedAvg+WSM and FedAvg when the number of participating clients is low. This
feature is significant since as explained above, low client participation is a known feature of
real world federated learning settings. We hypothesize the performance gap as a function of client fraction between the FedAvg and FedAvg+WSM is due to the larger impact of local client forgetting when we limit communication capability. Unless we actively take steps to control forgetting, non-participating clients will have their data distributions forgotten because unlike participating clients, they will be unable to contribute their updates to the global model. As the fraction of clients selected at each round increases, we observe the performance gap between the two methods narrow since more clients will have the opportunity to be selected at each round and "remind" the model of their data distributions.

\vspace{-8pt}\paragraph{Local Iterations} A local iteration is defined as one gradient step on the client during a federated round. In our reference setup $~7$ local iterations is equivalent to one local epoch. After 21 local iterations we observe a sharp decrease in accuracy for FedAvg as local iterations increase. While FedAvg+WSM does also experience a drop in accuracy as local iterations increase, its drop in accuracy is much less pronounced. This result is in line with what we would expect: since WSM helps to alleviate the effects of local client forgetting increasing the number of local training iterations will cause the effects of local client forgetting become more serious. The effect of WSM in allowing the local models to train for more local iterations without forgetting has important implications for the communication costs in FL. Since communication is a serious bottleneck and is limited by client drift, our method offers a valuable option to speed up training in a federated setting.  

\vspace{-8pt}\paragraph{Architecture}
We additionally investigate the performance of WSM using the LeNet architecture \citep{lecun1998gradient} on CIFAR-10 and CIFAR-100. While this model is not the considered state-of-the-art on these CIFAR datasets, it allows us to eliminate the dependence on the normalization as a hyperparameter to investigate relative performance for the purposes of this investigation. As before we train for 4000 rounds with each client training for three local epochs. The data is divided between clients using a Dirichlet distribution parameterized by $\alpha=0.1$. The reported values are the result of the average of three different seeds with the standard deviation indicating the variation between runs. Experiments were performed across a range of learning rates and the results of the best performing models with and without WSM are shown in Table~\ref{tab:lenet}. The complete set of results across all learning rates is provided in the appendix, section~\ref{sec:lenet_plus}. The difference between the best performing models of FedAvg and and FedAvg+WSM is $1.02\%$ for CIFAR-10 and an impressive $7.6\%$ on CIFAR-100. 

\begin{table*}[t]
\begin{center}
\caption{Accuracy results of FedAvg with and without WSM for different settings of client learning rates using the LeNet architecture.\vspace{-17pt}}
\label{tab:lenet}
\scriptsize
\begin{tabular}{ ccc } 
\toprule
 &CIFAR-10 &CIFAR-100\\ 
 \midrule
\textsc{  FedAvg} & $0.608\pm0.010$& $0.198\pm0.004$\\
\textsc{ FedAvg+WSM (ours)} &$\mathbf{0.624\pm0.009}$& $\mathbf{0.274\pm0.007}$\\ 
\bottomrule
\end{tabular}
\vspace{-5pt}
\end{center} 
\end{table*}

\vspace{-7pt}\section{Conclusion}\vspace{-7pt}
In this paper we took a deeper look at the \textit{local client forgetting} problem, we show that when a client performs local updates during FL, it risks overly optimizing its local objective leading to forgetting on other subsets of data. Local client forgetting degrades the performance of the global model and we show this phenomenon is especially severe in cases where there is a significant distribution mismatch across clients. First making the connection with the catastrophic forgetting problem in the continual learning, we propose a client level modification of the objective function which we call the weighted softmax. We show empirically that WSM allows us to mitigate client level forgetting by demonstrating improved performance for the FedAvg algorithm when combined with WSM across a range of learning rates.  We also demonstrate these improvements are not limited to FedAvg since we also observe significant performance increases when WSM is applied to SCAFFOLD and FedProx. An ablation study demonstrated WSM is particularly effective in the regime of highly heterogeneous client datasets and/or when a small percentage of clients are selected at each round. Our results indicate that addressing local client forgetting in general is an important consideration for federated learning optimization, one that bears closer scrutiny.

\bibliography{collas2023_conference}
\bibliographystyle{collas2023_conference}

\appendix
\section{Appendix}

\subsection{Additional Forgetting studies} \label{sec:additional_forgetting}
\begin{figure*}[t]
   \centering
  \includegraphics[width=.33\textwidth]{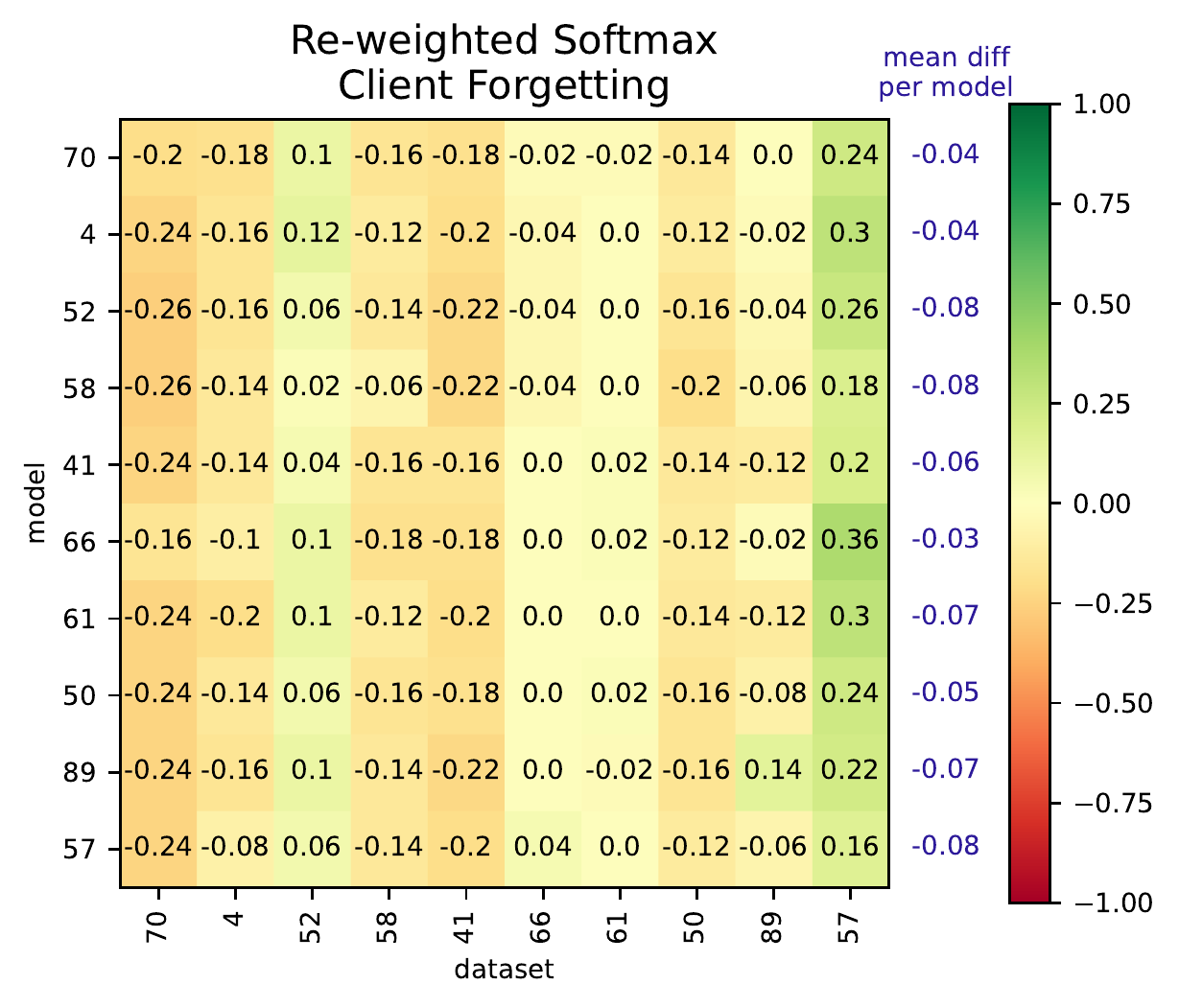}%round 1200 diff
  \includegraphics[width=.33\textwidth]{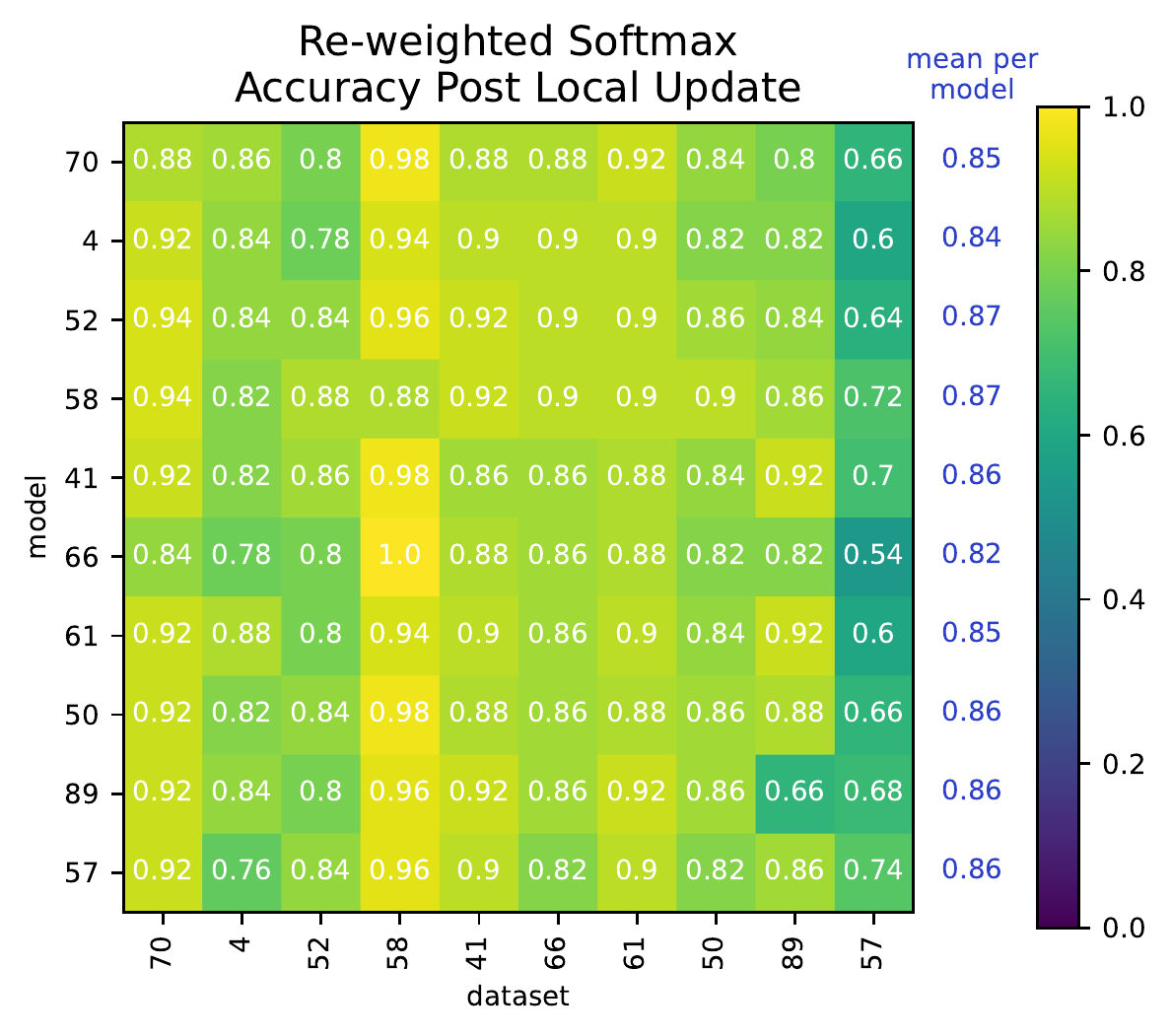}\\%round 1200 post
  \includegraphics[width=.33\textwidth]{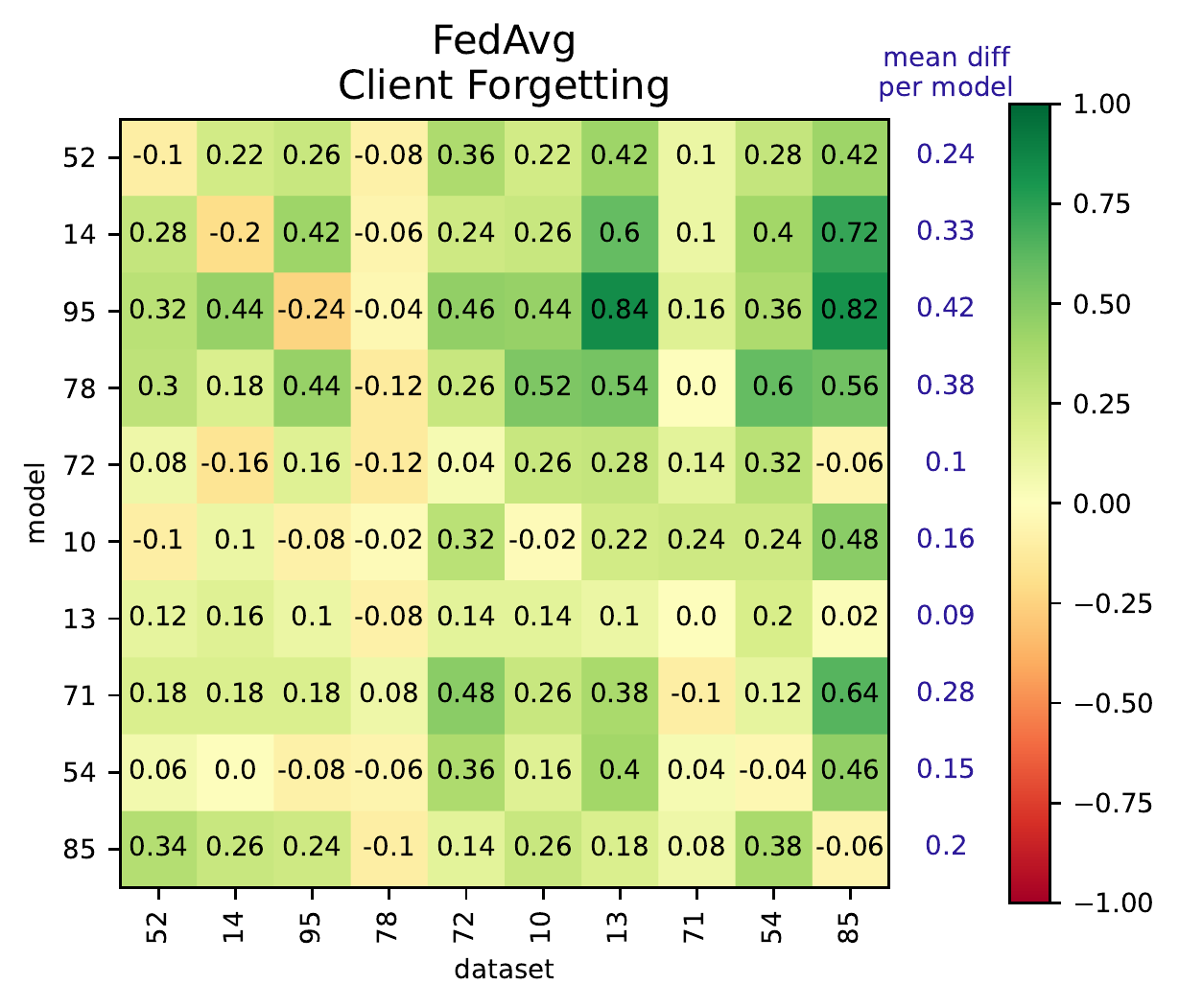}%round 1200 diff
  \includegraphics[width=.33\textwidth]{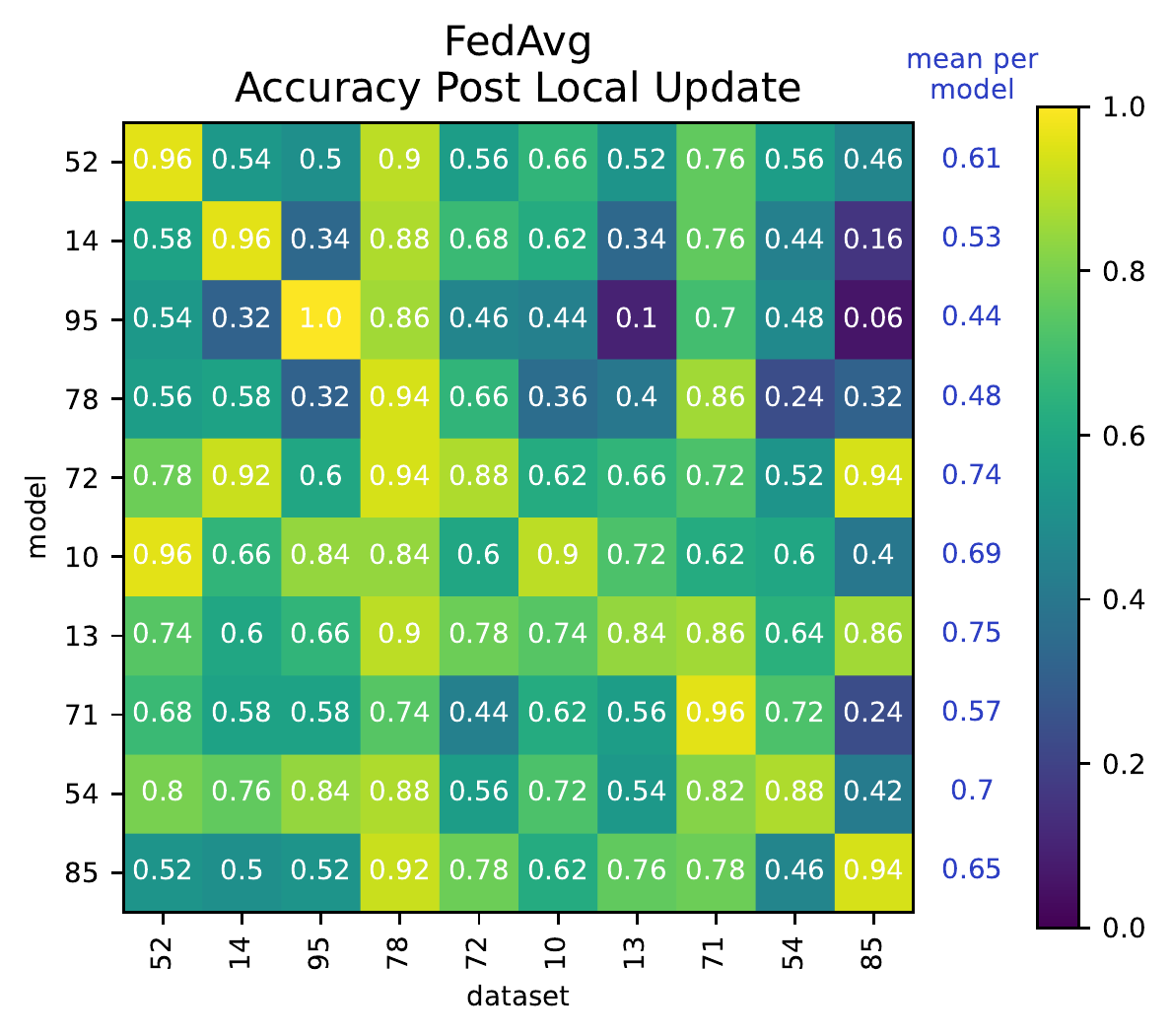}%round 1200 post
  \caption{We show local client forgetting for round 3600 with and without WSM. The heatmap is structured the same way as for figure~\ref{fig:client_forget} described previously where the y-axis indicates the model of client i and the x-axis indicates the local data of client k. Again at round 3600, much later in training than round 1200 shown in figure~\ref{fig:client_forget} we observe FedAvg+WSM (top row) significantly reduces forgetting across clients. \vspace{-10pt}} 
 \label{fig:client_forget3600}
\end{figure*}

In this section we show an additional heatmaps, similar to those from figure~\ref{fig:client_forget}. The heatmaps shown in figure~\ref{fig:client_forget} is for round 1200, $\frac{3}{10}$ of the way through training. Figure~\ref{fig:client_forget3600} shows heatmaps for round 3600, $\frac{9}{10}$ of the way through training which illustrates local client forgetting and the ability of WSM to reduce its effects are present throughout the training cycle.

\subsection{Complete Result Set Across Learning Rates and Normalization Methods }\label{sec:main_eval_plus}
Table~\ref{tab:main_extension} shows the complete set of results across learning rates and normalization methods. We observe FedAvg+WSM has a strong performance with batch normalization contrary to the findings of \citet{hsieh2020non} while FedAvg preforms better using group norm. 

\begin{table*}[t]
\caption{Accuracy results of FedAvg with and without WSM for different hyperparameters. We observe that FedAvg+WSM with batch normalization consistently improves performance over FedAvg, as well as having the highest overall accuracy by a large margin. WSM also makes the learning rate easier to tune since we observe a large hyperparameter range }
\small
\begin{center}
\begin{tabular}{ cccccc } 

\toprule
\multirow{ 2}{*}{Method}        & \multicolumn{2}{c}{\emph{Hyper-params}} & \multicolumn{3}{c}{\emph{\ \ \ \ \ \ \ Dataset}} \\
 &lr &norm &CIFAR-10 &CIFAR-100 &FEMNIST \\ 
 \midrule
\textsc{ FedAvg } & \multirow{ 4}{*}{0.5} & \multirow{ 2}{*}{group} &$0.326\pm 0.098$  & $0.292\pm 0.012$ & $0.542\pm 0.087$ \\
\textsc{ FedAvg+WSM (ours)} & & & $0.452\pm 0.173$ & $0.234\pm 0.191$ & $0.418\pm 0.177$\\
\textsc{ FedAvg } &  & \multirow{ 2}{*}{batch} &$0.742\pm 0.004$ &  $0.386\pm 0.003$ & $0.812\pm 0.020$ \\
\textsc{ FedAvg+WSM (ours)}& & &$\mathbf{0.792\pm 0.006}$& $\mathbf{0.426\pm 0.003}$& $\mathbf{0.837\pm 0.002}$\\
\cdashline{1-6}
\textsc{ FedAvg } & \multirow{ 4}{*}{0.3} & \multirow{ 2}{*}{group} &$0.791\pm 0.013$  & $0.384\pm 0.013$ & $0.769\pm 0.006$ \\
\textsc{ FedAvg+WSM (ours)} & & & $0.744\pm 0.007$ & $0.388\pm 0.027$ & $0.761\pm 0.118$\\
\textsc{ FedAvg } &  & \multirow{ 2}{*}{batch} &$0.742\pm 0.004$ &  $0.412\pm 0.014$ & $0.815\pm 0.008$ \\
\textsc{ FedAvg+WSM (ours)}& & &$\mathbf{0.834\pm0.008}$ &$\mathbf{0.467\pm0.015}$& $\mathbf{0.844\pm 0.010}$\\
\cdashline{1-6}
\textsc{ FedAvg } & \multirow{ 4}{*}{0.1} & \multirow{ 2}{*}{group} &$0.724\pm 0.027$  & \fcolorbox{red}{white}{$0.500\pm 0.016$} & $0.835\pm 0.002$ \\
\textsc{ FedAvg+WSM (ours)} & & & $0.794\pm 0.022$ & $0.446\pm 0.004$ & $0.827\pm 0.012$\\
\textsc{ FedAvg } &  & \multirow{ 2}{*}{batch} &\fcolorbox{orange}{white}{$0.820\pm 0.006$} &  $0.442\pm 0.016$ & $0.806\pm 0.031$\\
\textsc{ FedAvg+WSM (ours)}& & &$\mathbf{0.855\pm0.004}$ &$\mathbf{0.514\pm0.009}$& \fcolorbox{green}{white}{$\mathbf{0.848\pm 0.006}$}\\
\cdashline{1-6}
\textsc{ FedAvg } & \multirow{ 4}{*}{0.07} & \multirow{ 2}{*}{group} &$0.826\pm 0.007$  & $0.437\pm 0.007$ &  $\mathbf{0.827\pm 0.006}$  \\
\textsc{ FedAvg+WSM (ours)} & & & \fcolorbox{blue}{white}{$0.805\pm 0.007$} & \fcolorbox{blue}{white}{$0.484\pm 0.041$} & $0.823\pm 0.006$\\
\textsc{ FedAvg } &  & \multirow{ 2}{*}{batch} &$0.787\pm 0.006$ &  $0.513\pm 0.006$ & $0.789\pm 0.003$ \\
\textsc{ FedAvg+WSM (ours)}& & &$\mathbf{0.856\pm0.005}$ &$\mathbf{0.553\pm0.018}$& $0.826\pm 0.019$\\
\cdashline{1-6}
\textsc{ FedAvg } & \multirow{ 4}{*}{0.05} & \multirow{ 2}{*}{group} &$0.827\pm 0.004$  & $0.464\pm 0.001$ & \fcolorbox{red}{white}{$\mathbf{0.853\pm 0.004}$} \\
\textsc{ FedAvg+WSM (ours)} & & & $0.791\pm 0.019$ & $0.454\pm 0.015$ & \fcolorbox{blue}{white}{$0.841\pm 0.002$}\\
\textsc{ FedAvg } &  & \multirow{ 2}{*}{batch} &$0.790\pm 0.012$ &  $0.531\pm 0.007$ & \fcolorbox{orange}{white}{$0.833\pm 0.024$} \\
\textsc{ FedAvg+WSM (ours)}& & &\fcolorbox{green}{white}{$\mathbf{0.858\pm0.003}$} &$\mathbf{0.564\pm0.007}$& $0.842\pm 0.005$\\
\cdashline{1-6}
\textsc{ FedAvg } & \multirow{ 4}{*}{0.03} & \multirow{ 2}{*}{group} &\fcolorbox{red}{white}{$0.836\pm 0.005$} & $0.431\pm 0.020$ &  $\mathbf{0.835\pm0.006}$\\
\textsc{ FedAvg+WSM (ours)} & & & $0.774\pm 0.035$ & $0.472\pm 0.007$ & $0.830\pm 0.006$\\
\textsc{ FedAvg } &  & \multirow{ 2}{*}{batch} &$0.779\pm 0.028$ &  $0.561\pm 0.010$ & $0.756\pm0.015$ \\
\textsc{ FedAvg+WSM (ours)}& & &$\mathbf{0.857\pm0.005}$ &\fcolorbox{green}{white}{$\mathbf{0.581\pm0.005}$}&$0.834\pm 0.003$\\
\cdashline{1-6}
\textsc{ FedAvg } & \multirow{ 4}{*}{0.01} & \multirow{ 2}{*}{group} &$0.815\pm 0.003$  & $0.431\pm 0.005$ & $\mathbf{0.830\pm0.002}$ \\
\textsc{ FedAvg+WSM (ours)} & & & $0.785\pm 0.011$ & $0.471\pm 0.010$ & $0.800\pm 0.018$\\
\textsc{ FedAvg } &  & \multirow{ 2}{*}{batch} &$0.787\pm 0.003$ &  $0.566\pm 0.009$ & $0.744\pm0.011$ \\
\textsc{ FedAvg+WSM (ours)}& & &$\mathbf{0.845\pm0.006}$ &$\mathbf{0.574\pm0.006}$& $0.800\pm0.019$\\
\cdashline{1-6}
\textsc{ FedAvg } & \multirow{ 4}{*}{0.007} & \multirow{ 2}{*}{group} &$0.817\pm 0.007$  & $0.426\pm 0.005$ & $\mathbf{0.821\pm 0.011}$ \\
\textsc{ FedAvg+WSM (ours)} & & & $0.773\pm 0.014$ & $0.476\pm 0.010$ & $0.794\pm 0.006$\\
\textsc{ FedAvg } &  & \multirow{ 2}{*}{batch} &$0.797\pm 0.008$ &  \fcolorbox{orange}{white}{$\mathbf{0.568\pm0.003}$} & $0.734\pm0.018$ \\
\textsc{ FedAvg+WSM (ours)}& & &$\mathbf{0.841\pm0.004}$ &$\mathbf{0.568\pm0.003}$& $0.800\pm0.007$ \\
\cdashline{1-6}
\textsc{ FedAvg } & \multirow{ 4}{*}{0.005} & \multirow{ 2}{*}{group} &$0.802\pm 0.010$  & $0.426\pm 0.002$ & $\mathbf{0.819\pm0.022}$ \\
\textsc{ FedAvg+WSM (ours)} & & & $0.768\pm 0.019$ & $0.474\pm 0.006$ & $0.781\pm0.017$\\
\textsc{ FedAvg } &  & \multirow{ 2}{*}{batch} &$0.783\pm 0.009$ &  $0.553\pm 0.005$ & $0.743\pm0.053$  \\
\textsc{ FedAvg+WSM (ours)}& & &$\mathbf{0.826\pm0.009}$ &$\mathbf{0.554\pm0.004}$& $0.773\pm0.013$\\

\bottomrule
\end{tabular}
\end{center}
\label{tab:main_extension}
\end{table*}

\subsection{LeNet Performance Across Multiple Learning Rates} \label{sec:lenet_plus}
Table~\ref{tab:lenet_all} shows model performance using the LeNet architecture across learning rates. WSM outperforms vanilla FedAvg for each learning rate for both datasets. As with the ResNet-18 case, we continue to observe that WSM provides good performance over a larger range of learning rates than FedAvg which makes it easier to tune. We also observe WSM reduces the variance in model accuracy as evidenced by the typically lower standard deviations reported for the runs in each set.

\begin{table*}[t]
\begin{center}
\caption{Accuracy results of FedAvg with and without WSM for different settings of client learning rates using the LeNet architecture.\vspace{-12pt}} 
\label{tab:lenet_all}
\begin{tabular}{ ccccc } 
\toprule
 &lr &CIFAR-10 &CIFAR-100\\ 
 \midrule
\textsc{ FedAvg } & &$0.589\pm0.012$ &$0.119\pm0.002$ \\
\textsc{ FedAvg+WSM (ours)}& 0.07 &$\mathbf{0.624\pm0.009}$ &$\mathbf{0.180\pm0.013}$ \\\cdashline{1-5}
\textsc{ FedAvg} & &$0.580\pm0.051$ &$0.154\pm0.011$ \\
\textsc{ FedAvg+WSM (ours)} &0.05 &$\mathbf{0.615\pm0.004}$&$\mathbf{0.217\pm0.009}$ \\\cdashline{1-5}
\textsc{ FedAvg} & &$0.605\pm0.014$ &$0.168\pm0.001$ \\
\textsc{ FedAvg+WSM (ours)} &0.03 &$\mathbf{0.621\pm0.009}$ &$\mathbf{0.251\pm0.007}$ \\\cdashline{1-5}
\textsc{  FedAvg} & &\fcolorbox{red}{white}{$0.608\pm0.010$}&$0.168\pm0.044$ \\
\textsc{ FedAvg+WSM (ours)} &0.01 &\fcolorbox{green}{white}{$\mathbf{0.624\pm0.009}$}&\fcolorbox{green}{white}{$\mathbf{0.274\pm0.007}$} \\\cdashline{1-5}
\textsc{ FedAvg} & &$0.591\pm0.014$ &$0.195\pm0.004$ \\
\textsc{ FedAvg+WSM (ours)} &0.007&$\mathbf{0.620\pm0.022}$&\fcolorbox{green}{white}{$\mathbf{0.274\pm0.001}$} \\\cdashline{1-5}
\textsc{ FedAvg} & & $0.585\pm0.019$&\fcolorbox{red}{white}{$0.198\pm0.004$} \\
\textsc{ FedAvg+WSM (ours)} &0.005& $\mathbf{0.615\pm0.003}$&$\mathbf{0.270\pm0.001}$ \\\cdashline{1-5}
\textsc{ FedAvg}  & &$0.545\pm0.018$ &$0.181\pm0.009$\\
\textsc{ FedAvg+WSM (ours)} &0.003& $\mathbf{0.566\pm0.001}$ &$\mathbf{0.247\pm0.004}$ \\\cdashline{1-5}
\textsc{ FedAvg} & &$0.441\pm0.019$ &$0.115\pm0.007$ \\
\textsc{ FedAvg+WSM (ours)} &0.001&$\mathbf{0.478\pm0.009}$ &$\mathbf{0.186\pm0.003}$\\\cdashline{1-5}
\bottomrule
\end{tabular}
\end{center}
\end{table*}

\subsection{Illustrating different dirichlet parameters}\label{sec:dir_alphas}
\begin{figure*}[t]
    \centering
    \includegraphics[width=.29\textwidth]{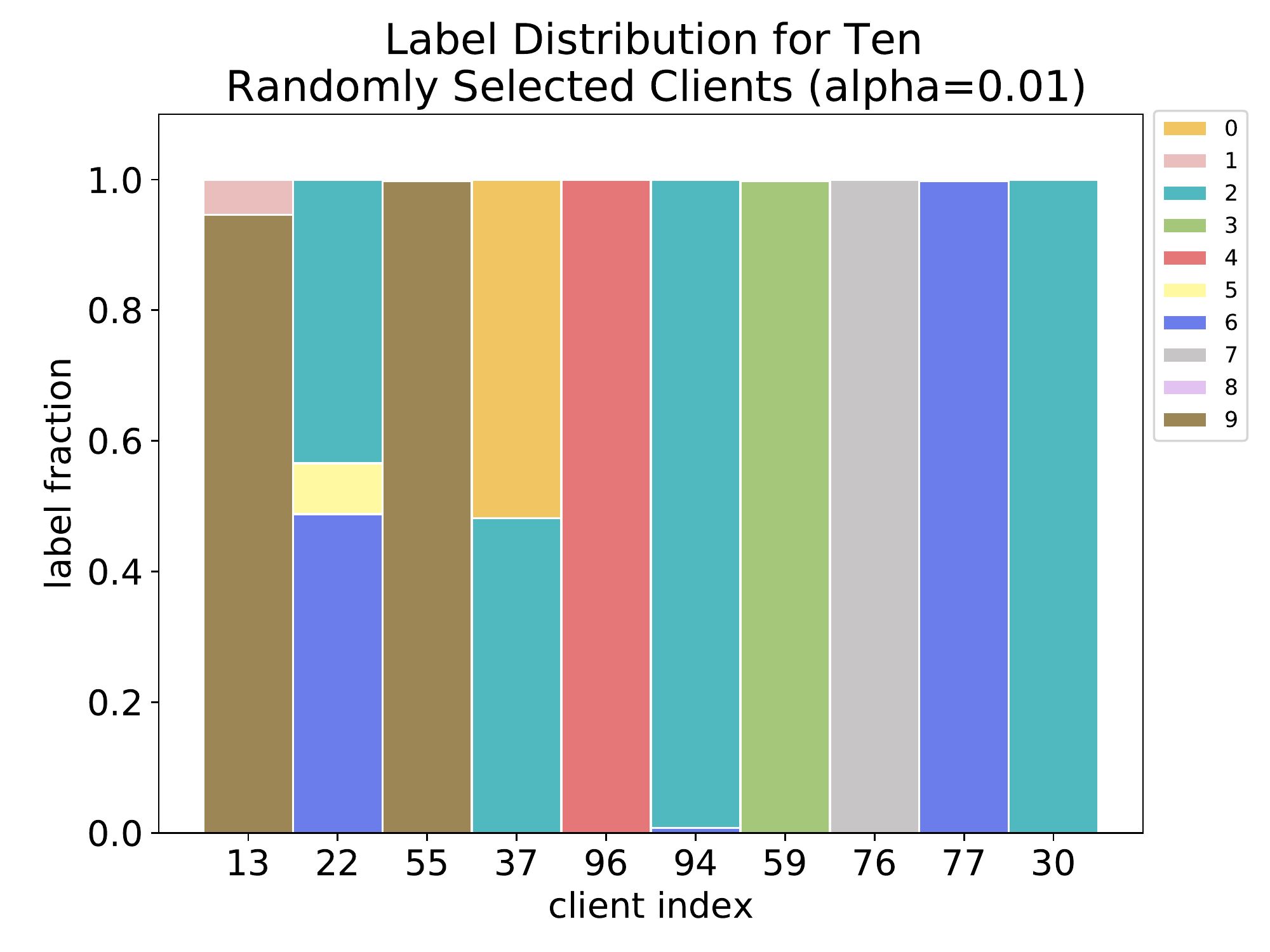}
    \includegraphics[width=.29\textwidth]{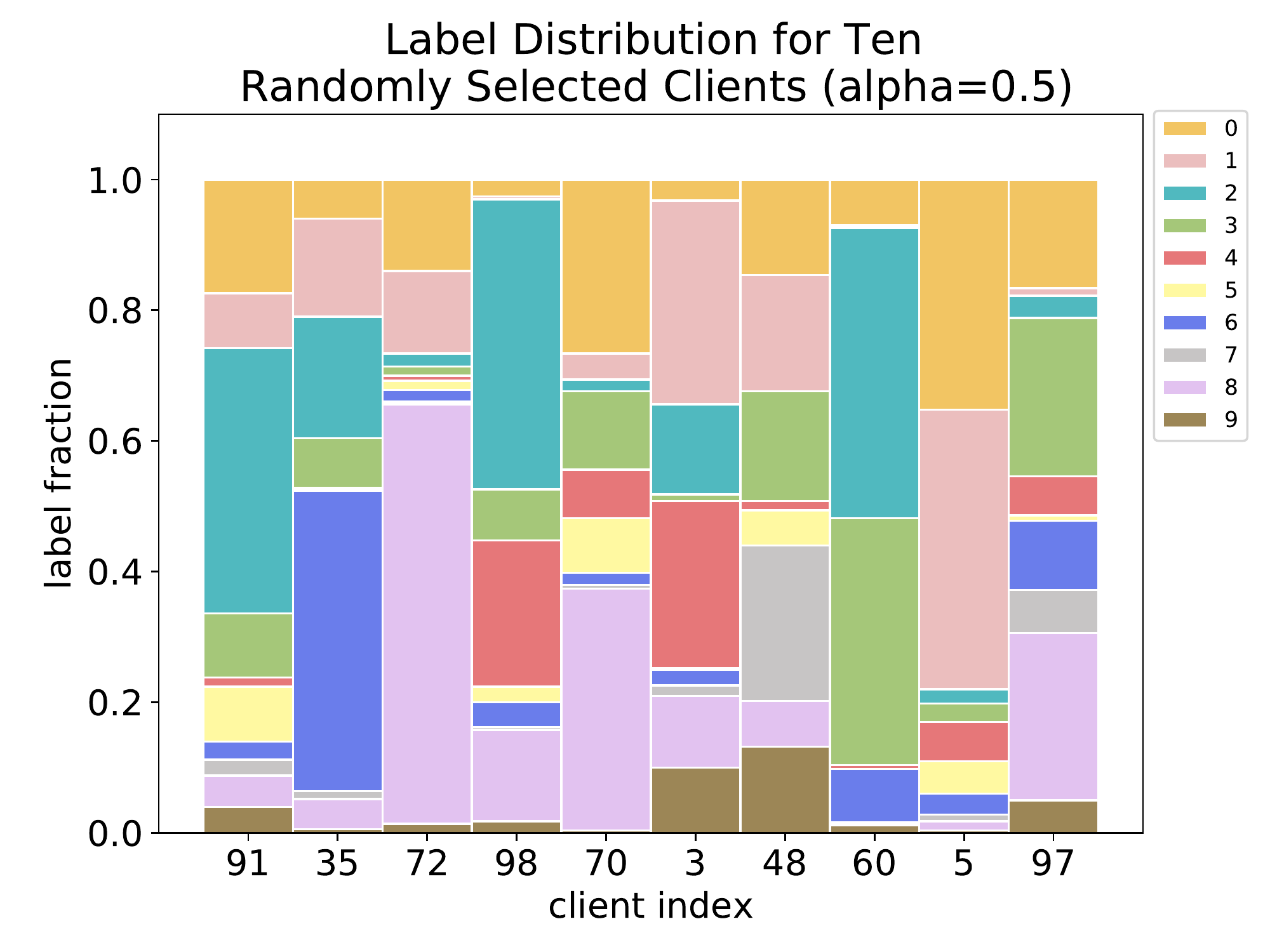}
    \includegraphics[width=.29\textwidth]{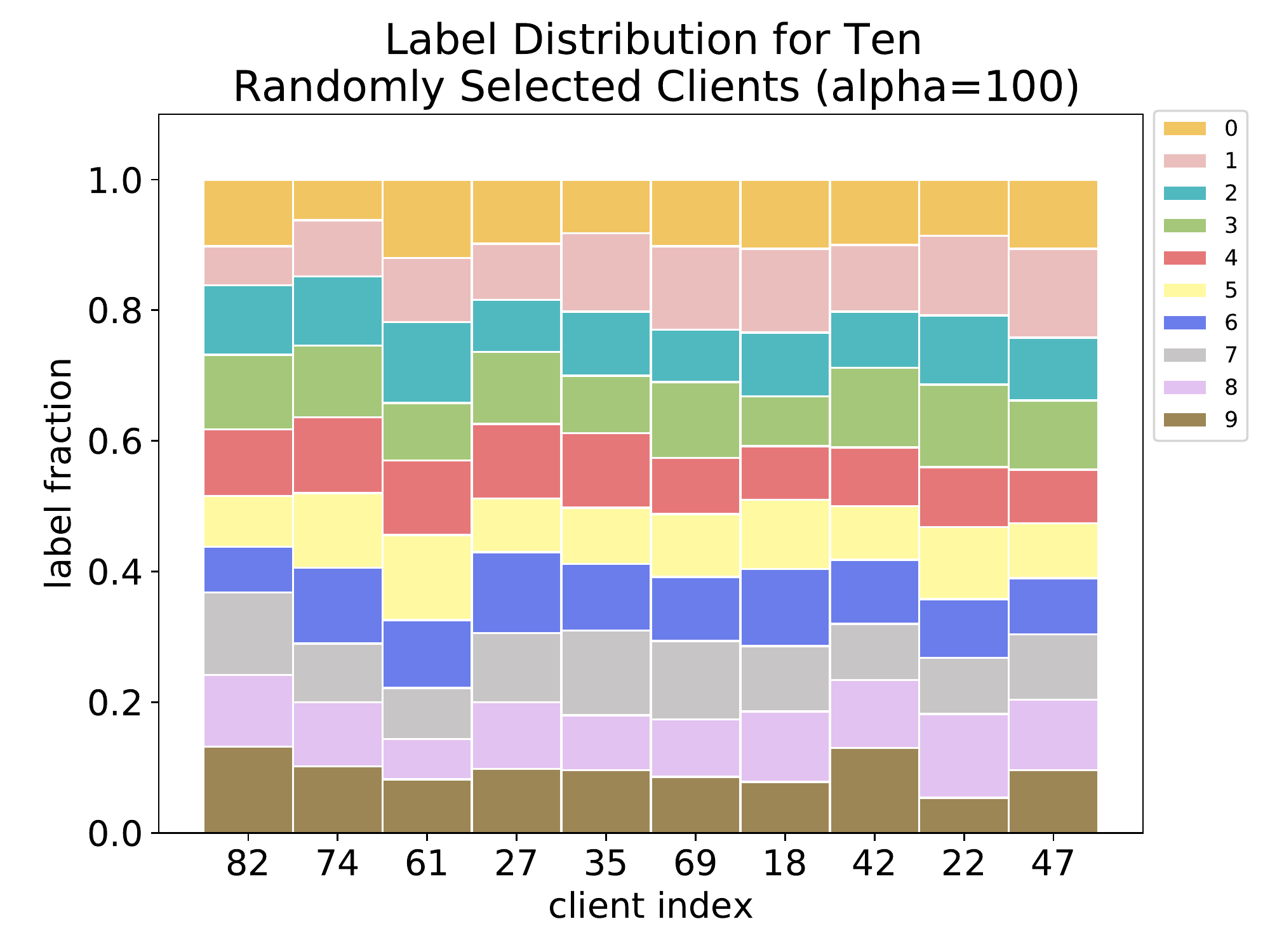}
    \caption{Percentages of each class label for ten randomly selected clients with $\alpha=0.01, 0.5, 100$ from left to right} 
    \label{fig:alphas}
\end{figure*}

Figure \ref{fig:alphas} offers a practical illustration of how client partitions change as a function of $\alpha$. Clients with $\alpha=0.01$ have only a small percentage of the classes while at $\alpha = 100$ clients have all 10 classes in proportions that are much more equal than we observe with the other two parameterizations.  

\end{document}